\documentclass[lettersize,journal]{IEEEtran}
\usepackage{amsmath,amsfonts}
\usepackage{algorithmic}
\usepackage{algorithm}
\usepackage{array}
\usepackage[caption=false,font=normalsize,labelfont=sf,textfont=sf]{subfig}
\usepackage{textcomp}
\usepackage{stfloats}
\usepackage{url}
\usepackage{verbatim}
\usepackage{graphicx}
\usepackage{cite}
\usepackage{xcolor}
\usepackage{makecell}
\usepackage{threeparttable}
\usepackage{hyperref}
\usepackage{ulem}

\begin{document}

\title{BIAS: A Biologically Inspired Algorithm for Video Saliency Detection}

\author{Zhao-ji Zhang, Ya-tang Li
\thanks{This work was supported in part by the Natural Science Foundation of Beijing Municipality under Grant IS23073, in part by the National Natural Science Foundation of China under Grant 32271060. \textit{(Corresponding authors: Ya-tang Li)}}
\thanks{Zhao-ji Zhang is with the Academy for Advanced Interdisciplinary Studies, Peking University, Beijing 100871, China, and also with Chinese Institute for Brain Research, Beijing (CIBR), Beijing 102206, China (email: zhangzhaoji@cibr.ac.cn)}
\thanks{Ya-tang Li is with the Chinese Institute for Brain Research, Beijing (CIBR), Beijing 102206, China (e-mail: yatangli@cibr.ac.cn).}
}
\markboth{}%
{Shell \MakeLowercase{\textit{Zhang Z and Li YT}}: A Biologically Inspired Algorithm for Video Saliency Detection}

\maketitle

\begin{abstract}
We present BIAS, a fast, biologically inspired model for dynamic visual saliency detection in continuous video streams. Building on the Itti--Koch framework, BIAS incorporates a retina-inspired motion detector to extract temporal features, enabling the generation of saliency maps that integrate both static and motion information. Foci of attention (FOAs) are identified using a greedy multi-Gaussian peak-fitting algorithm that balances winner-take-all competition with information maximization. BIAS detects salient regions with millisecond-scale latency and outperforms heuristic-based approaches and several deep-learning models on the DHF1K dataset, particularly in videos dominated by bottom-up attention. Applied to traffic accident analysis, BIAS demonstrates strong real-world utility, achieving state-of-the-art performance in cause-effect recognition and anticipating accidents up to 0.72 seconds before manual annotation with reliable accuracy. Overall, BIAS bridges biological plausibility and computational efficiency to achieve interpretable, high-speed dynamic saliency detection. 
\end{abstract}


\section{Introduction}
\label{sec:intro}
Each second, our retinas are bombarded with patterns of photons delivering roughly $10^9$ bits of visual information~\cite{kelly_information_1962, zhaoping_new_2019, koch_how_2006}, yet human perception can process only about 20 bits per second~\cite{zheng_unbearable_2025}.
This vast gap between sensory input and perceptual capacity poses a fundamental computational challenge: the brain must efficiently extract a small fraction of behaviorally relevant signals from an overwhelming, continuous sensory stream. Visual attention addresses this challenge by allocating limited processing resources to the most informative regions of the visual field.

Attention can be driven either involuntarily by salient external stimuli or voluntarily by internal goals~\cite{petersen_attention_2012}. To model bottom-up attention, Koch and Ullman, building on feature integration theory~\cite{treisman_feature-integration_1980}, introduced the concept of a \textit{saliency map}~\cite{koch_shifts_1985}. This map integrates multiple feature channels into a retinotopic representation of saliency strength, where a winner-take-all (WTA) network selects the most salient location. Itti et al.~\cite{itti_model_1998} implemented this model for static images, igniting decades of research spanning classical computer vision to modern deep learning approaches~\cite{borji_state---art_2013,itti_computational_2001}.

Most prior work focuses on \textit{static images}, capturing only a single moment in time, whereas natural vision is inherently dynamic, requiring continuous attention over changing inputs. While deep learning has advanced video saliency detection, existing models remain computationally demanding, often biologically implausible, and are unsuitable for real-time applications~\cite{wang_revisiting_2021}. These limitations are particularly critical in time-sensitive scenarios such as traffic accident anticipation—a leading cause of mortality worldwide~\cite{noauthor_global_2023}—where rapid visual inference is essential for human safety and autonomous driving~\cite{chan_anticipating_2017, yao_dota_2023, wang_revisiting_2018, kopuklu_driver_2021, you_traffic_2020, fang_dada-2000_2019}.

We introduce \textbf{BIAS} (Biologically Inspired Algorithm for video Saliency detection) for fast, interpretable, and spatiotemporal saliency prediction. Building on the Itti–Koch framework, BIAS integrates motion information via a retina-inspired detector and selects attended locations using a Gaussian Winner-Take-All (GWTA) mechanism, balancing efficiency, biological plausibility, and computational interpretability.

On the DHF1K benchmark~\cite{wang_revisiting_2018}, BIAS outperforms heuristic-based models and approaches the performance of modern deep networks, while achieving substantially faster runtime. When applied to the traffic accident benchmark for causality recognition~\cite{you_traffic_2020}, BIAS achieves state-of-the-art performance in cause–effect recognition and predicts collision causes an average of 0.72 s before manual annotation. These results demonstrate its practical utility in real-time, safety-critical scenarios and suggest that dynamic saliency plays an important role in accident detection.

\textbf{To summarize our contributions:}
\begin{enumerate}
    \item We propose \textit{BIAS}, a fast, interpretable, bio-inspired video saliency detection model that balances predictive performance and computational efficiency.
    \item We introduce a \textit{motion saliency detector} inspired by Hassenstein--Reichardt models, capturing both motion direction and speed.
    \item We accelerate the saliency computation using \textit{kernel-decomposed Gabor filtering}.
    \item We develop a \textit{Gaussian Winner-Take-All (GWTA)} method for robust and efficient fixation selection.
    \item We demonstrate BIAS's utility in \textit{traffic accident analysis}, where it enables causality recognition and reliable early anticipation.
\end{enumerate}

\section{Related Works}

\label{sec:related_work}

\subsection{Heuristic-based models for video saliency detection}


Compared to static image saliency, video saliency detection introduces additional complexity due to motion and temporal dynamics. \IEEEpubidadjcol Early motion saliency research primarily focused on surveillance applications~\cite{wildes_measure_1998,wixson_detecting_2000}. A major wave of subsequent work extended the Itti–Koch framework for static saliency~\cite{itti_model_1998,itti_computational_2001}, adapting local contrast-based mechanisms to capture motion cues and constructing spatiotemporal master saliency maps by integrating static and dynamic features~\cite{gao_discriminant_2007, ma_model_2002, seo_static_2009, kim_spatiotemporal_2011, mahadevan_spatiotemporal_2010, fang_video_2014, ma_generic_2005, zhai_visual_2006, le_meur_predicting_2007}.  

Beyond contrast-based methods, diverse strategies have been explored, including video compression optimization~\cite{khatoonabadi_how_2015,muthuswamy_salient_2013,sinha_region--interest_2004}, Bayesian inference~\cite{itti_bayesian_2009,zhang_sun_2008}, motion-energy modeling~\cite{ma_new_2001}, spectral analysis~\cite{hou_dynamic_2008,guo_spatio-temporal_2008,guo_novel_2010}, feature whitening~\cite{leboran_dynamic_2017,muthuswamy_salient_2013}, and self-resemblance-based measures~\cite{seo_static_2009}.  

However, these classical models typically show limited performance on modern large-scale benchmarks~\cite{wang_revisiting_2018} due to their reliance on hand-crafted features and lack of semantic understanding.

\subsection{Learning-based models for video saliency detection}

The availability of large-scale human fixation datasets~\cite{wang_revisiting_2018,mital_clustering_2011,mathe_actions_2015,hadizadeh_eye-tracking_2012,itti_automatic_2004} has fueled rapid progress in learning-based video saliency detection. Unlike purely bottom-up heuristic models, these methods combine both stimulus-driven and task-driven cues to predict attention.

Early deep models adopted two-stream convolutional architectures~\cite{simonyan_two-stream_2014}, inspired by the ventral and dorsal visual pathways. These architectures process spatial and temporal information separately before fusing them for spatiotemporal saliency estimation~\cite{jiang_deepvs_2018,bak_spatio-temporal_2018,zhang_video_2019}.  
To better capture long-range temporal dependencies, later works combined CNNs with recurrent units such as LSTMs, enabling temporal propagation across frames and improving consistency and benchmark performance~\cite{wang_revisiting_2018,wu_salsac_2020,linardos_simple_2019,gorji_going_2018,droste_unified_2020,cornia_predicting_2018,zhang_spatial-temporal_2021}.  
Alternatively, 3D CNNs process video volumes directly, providing richer temporal features and smoother attentional transitions~\cite{fang_deep3dsaliency_2018,min_tased-net_2019,wang_spatio-temporal_2023,bellitto_hierarchical_2021,lai_video_2019,chang_temporal-spatial_2021,jain_vinet_2021}.  

Recently, transformer-based architectures have become dominant~\cite{zhou_transformer-based_2023,moradi_salfom_2025,li_tm2sp_2025,wang_spatio-temporal_2023,ma_video_2022,moradi_transformer-based_2024}. Leveraging global self-attention, they effectively model long-range spatiotemporal dependencies and anticipate future gaze shifts. However, these models are computationally heavy and biologically opaque, posing challenges for real-time or resource-constrained applications such as robotics and autonomous driving.

\subsection{Traffic accident anticipation}

Traffic safety analysis has drawn increasing attention with the growth of autonomous driving datasets~\cite{sun_scalability_2020,ettinger_large_2021}. However, accident data exhibit severe long-tail statistics~\cite{codevilla_exploring_2019}, making supervised learning difficult due to the scarcity of annotated crash events. Even specialized datasets such as TUMTraf-A~\cite{zimmer_safety-critical_2025} remain limited in scale.  

To mitigate annotation scarcity, several dashboard-camera datasets have been developed for accident anticipation~\cite{chan_anticipating_2017,fang_dada-2000_2019,bao_uncertainty-based_2020,yao_dota_2023,you_traffic_2020,ali_advances_2024}. Representative approaches include Bayesian uncertainty modeling \cite{bao_uncertainty-based_2020}, semantic scene parsing~\cite{you_traffic_2020}, trajectory forecasting~\cite{Li_traffic_2025,li_prediction_2025}, and spatiotemporal graph neural networks~\cite{karim_dynamic_2022}. While these methods can predict the likelihood of a collision, they typically lack causal reasoning and cannot explicitly identify the cues that lead to accidents.  


\section{Approach}
\label{sec:approach}

We propose a biologically inspired algorithm for dynamic visual saliency detection in continuous video streams (Fig.~\ref{fig.1}a). The framework integrates static saliency from individual frames with motion-based saliency derived from temporal cues, combining both into a unified spatiotemporal saliency representation.

\subsection{Image saliency detection}

The static saliency computation extends the classical bottom-up model~\cite{itti_model_1998,peters_applying_2008} by improving efficiency and representational fidelity. Each video frame is represented as a tensor $\mathbf{I}\in\mathbb{R}^{H\times W\times 3}$ with resolution $480\times640$ and RGB channels.

\textbf{Intensity channels.}  
Two intensity channels capture luminance polarity:
{\small
\begin{align}
\mathbf{I_+} &= 0.299\mathbf{R} + 0.587\mathbf{G} + 0.114\mathbf{B}, \\
\mathbf{I_-} &= 255 - \mathbf{I_+},
\end{align}
}
following the ITU-R BT.601 standard~\cite{article_noauthor_ycbcr_2025}.
Each channel is Gaussian-blurred and progressively down-sampled by a factor of two to construct pyramids $\mathbf{I_+}(\sigma)$ and $\mathbf{I_-}(\sigma)$ for $\sigma \in [0,8]$.

\textbf{Color channels.}  
Four opponent color channels are computed as:
{\small
\begin{align}
\mathbf{\tilde{R}} &= \mathbf{R} - \tfrac{\mathbf{G}+\mathbf{B}}{2}, \\
\mathbf{\tilde{G}} &= \mathbf{G} - \tfrac{\mathbf{R}+\mathbf{B}}{2}, \\
\mathbf{\tilde{B}} &= \mathbf{B} - \tfrac{\mathbf{R}+\mathbf{G}}{2}, \\
\mathbf{\tilde{Y}} &= \tfrac{\mathbf{R}+\mathbf{G}}{2} - \tfrac{|\mathbf{R}-\mathbf{G}|}{2} - \mathbf{B}.
\end{align}
}
Gaussian pyramids are then constructed for each channel.

\textbf{Orientation channels.}  
In the original model, four orientation channels $\mathbf{O}(\sigma,\theta)$ are obtained by convolving the intensity map $I$ with 2D Gabor filters at $\theta\in{0,\pi/4,\pi/2,3\pi/4}$, at a cost of $\mathcal{O}(D^2HW)$. To reduce computation, we employ a separable kernel decomposition that reduces the complexity to $\mathcal{O}(DHW)$. For any 2D function $f(x,y)$,
{\small
\begin{align}
&(f * F_{\omega, \theta, \sigma})(x,y) \notag \\
&= \iint f(x,y) \exp{\left(-\frac{(x-k)^2+(y-l)^2}{2\sigma^2}\right)} \notag \\
  &\hspace{3em} \exp{\left(\frac{2\pi \omega i\big[(x-k)\cos\theta + (y-l)\sin\theta)\big]}{\lambda}\right)} \mathrm{d}k\mathrm{d}l \notag \\
  &= \iint f(x,y) \exp{\left(-\frac{(x-k)^2}{2\sigma^2}\right)}\exp{\left(-\frac{(y-l)^2}{2\sigma^2}\right)}\notag \\
  &\quad \exp{\left(\frac{2\pi \omega i(x-k)\cos\theta}{\lambda}\right)} \exp{\left(\frac{2\pi \omega i(y-l)\sin\theta}{\lambda}\right)} \mathrm{d}k\mathrm{d}l \notag \\
    &= \iint f(x,y) G_{\sigma}\left(x-k\right)G_{\sigma}\left(y-l\right) \notag \\ 
    &\hspace{3em} H_{\omega,\theta}\left(x-k\right)V_{\omega,\theta}\left(y-l\right) \mathrm{d}k\mathrm{d}l
\end{align}
}
Here $F_{\omega,\theta,\sigma_F}$ is a 2D Gabor with spatial frequency $\omega$, orientation $\theta$, and Gaussian envelope $\sigma_F$, where we set $\sigma_F=2\pi^2/\omega$~\cite{kim_fast_2018}. For pyramid levels with $\sigma<5$, we use $\omega=2\pi^2/2.7$; for $\sigma\ge5$, we reuse the $\sigma=5$ features while doubling the wavelength, effectively detecting subsampled high-level features with reduced variability. This accelerated scheme yields four orientation pyramids $\mathbf{O}(\sigma,\theta)$ per image at substantially lower computational cost.

\begin{figure}[tb!]
    \centering
    \subfloat[]{\includegraphics[width=1\linewidth]{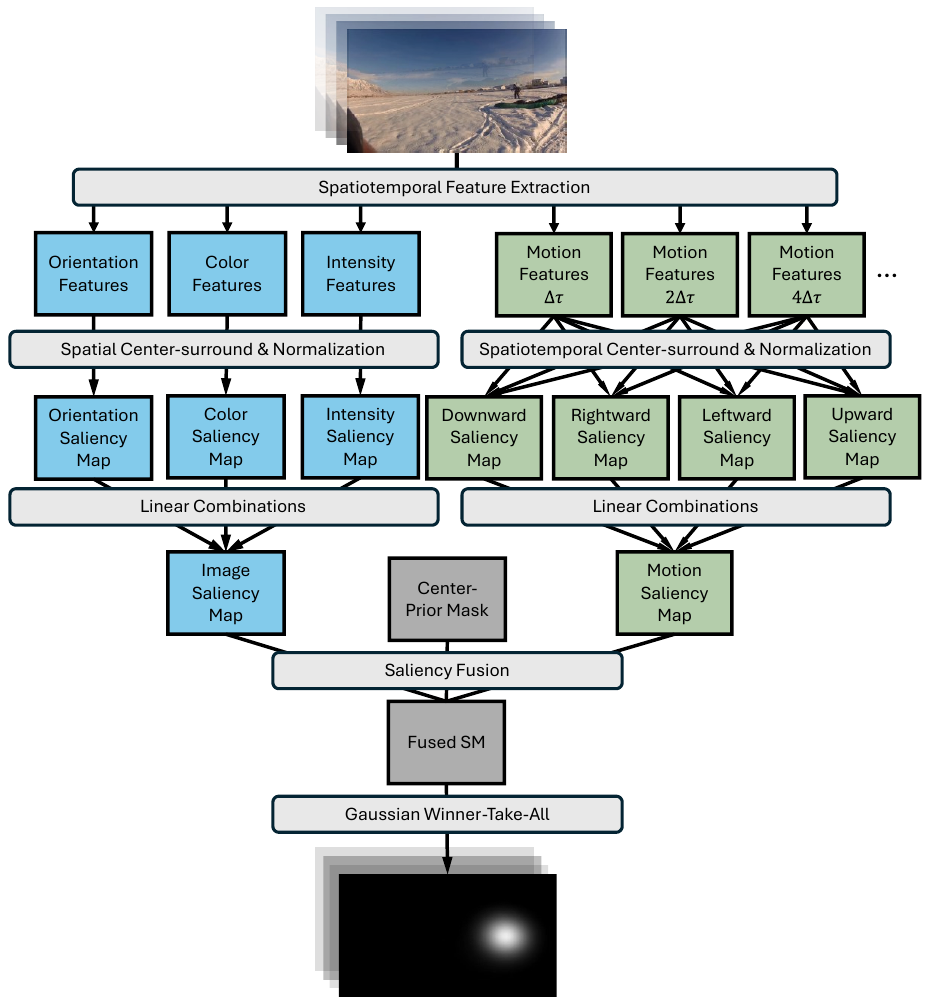}}
    \newline
    \subfloat[]{\includegraphics[width=0.55\linewidth]{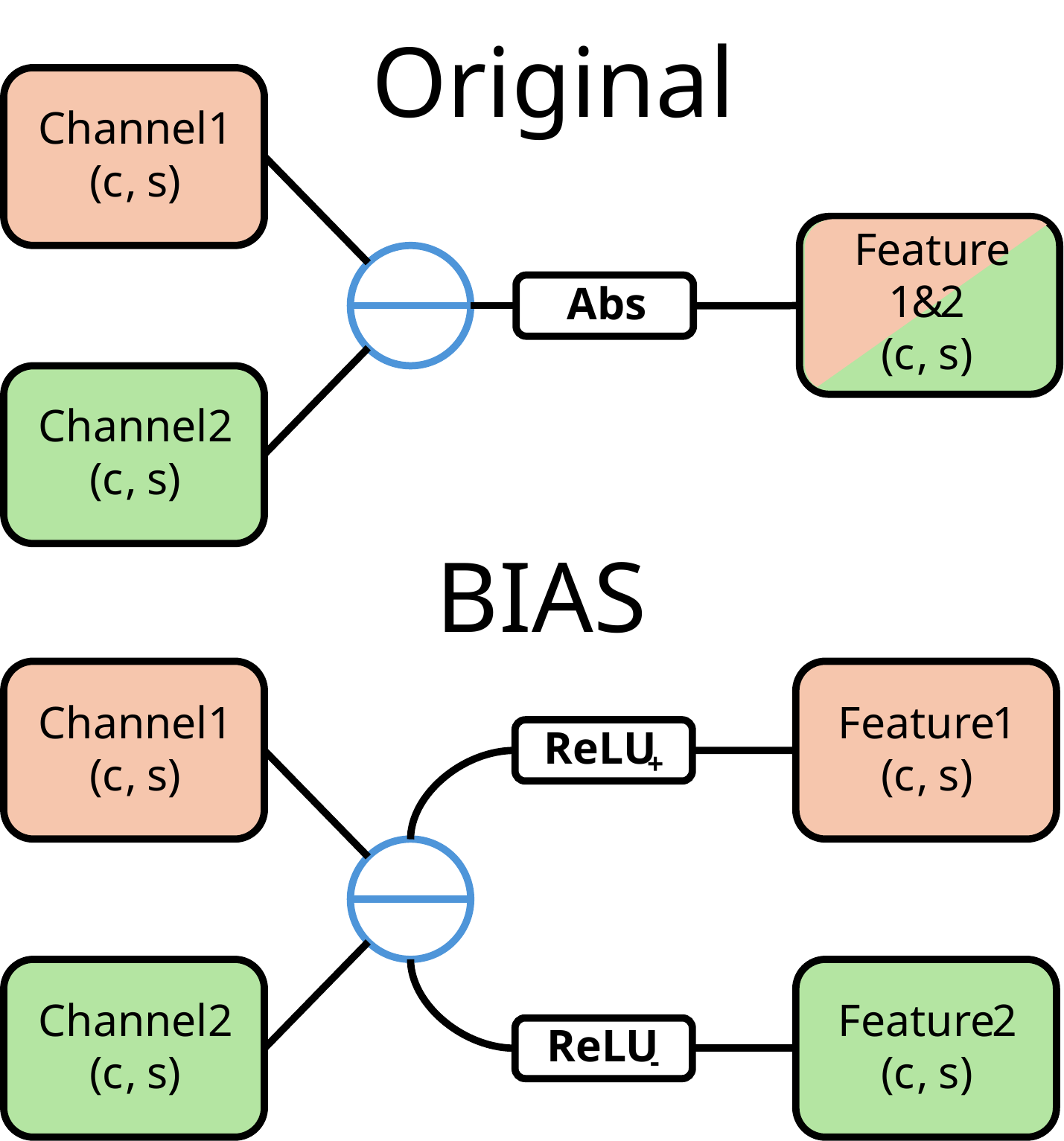}}
    \subfloat[]{\includegraphics[width=0.39\linewidth]{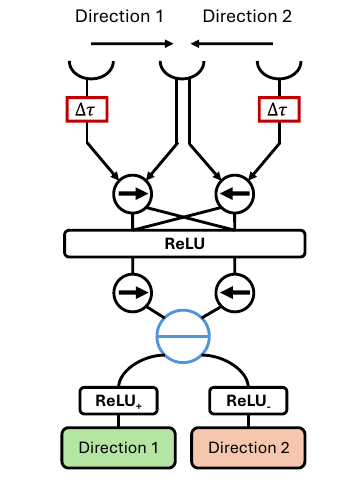}}
    \caption{
    (a) General architecture of BIAS.
    (b) Comparison of center–surround computation between Itti's original model (top) and BIAS (bottom).
    (c) Motion computation inspired by direction-selective cells in the fly retina.
    }
    \label{fig.1}
\end{figure}

\textbf{Feature maps.}  
Feature maps are computed via center–surround operations $(\ominus)$ across pyramid levels. Unlike the original model, we adopt ReLU-based half-wave rectification to preserve opponent polarity (Fig.~\ref{fig.1}b):
{\small
\begin{align}
    \mathbf{F_{+}}(c,s) &= \text{ReLU}\!\Big((\mathbf{F_{+}}(c)-\mathbf{F_{-}}(c)) \ominus (\mathbf{F_{-}}(s)-\mathbf{F_{+}}(s)\Big) \\
    \mathbf{F_{-}}(c,s) &= \text{ReLU}\!\Big((\mathbf{F_{-}}(c)-\mathbf{F_{+}}(c)) \ominus (\mathbf{F_{+}}(s)-\mathbf{F_{-}}(s)\Big)
\end{align}
}
Here, $(\mathbf{F_+},\mathbf{F_-})$ denotes opponent feature pairs such as $(\mathbf{I_+},\mathbf{I_-})$ or $(\mathbf{R},\mathbf{G})$.
Orientation feature maps are computed as:
{\small
\begin{equation}
\mathbf{O}(c,s,\theta)=\big|\mathbf{O}(c,\theta)\ominus\mathbf{O}(s,\theta)\big|.
\end{equation}
}
Using $c\in[2,3,4]$ and $s=c+\delta$ for $\delta\in[1,2,3,4]$, this yields 120 feature maps (24 intensity, 48 color, 48 orientation). 




\textbf{Normalization and conspicuity maps.}  
Each feature map is normalized using
{\small
\begin{equation}
    \mathcal{N}(\mathbf{X}) = (M-\bar m)^2 \frac{\mathbf{X}}{M},
\end{equation}
}
where $M$ and $\bar{m}$ denote global and local maxima, respectively.

Conspicuity maps for intensity ($\mathbf{\bar I}$), color ($\mathbf{\bar C}$), and orientation ($\mathbf{\bar O}$) are obtained by summing normalized feature maps across scales and channels:
{\small
\begin{align}
    \mathbf{\bar I} &= \bigoplus_{c}\bigoplus_{s=c+\delta}\Big(\mathcal{N}\big(\mathbf{I_+}(c,s)\big)+\mathcal{N}\big(\mathbf{I_-}(c,s)\big)\Big) \\
    \mathbf{\bar C} &= \bigoplus_{c}\bigoplus_{s=c+\delta}\bigg(\mathcal{N}\big(\mathbf{RG_+}(c,s)\big)+\mathcal{N}\big(\mathbf{RG_-}(c,s)\big) \notag \\
    &\quad +\mathcal{N}\big(\mathbf{BY_+}(c,s)\big)+\mathcal{N}\big(\mathbf{BY_-}(c,s)\big)\bigg) \\
    \mathbf{\bar O} &= \sum_{\theta \in \{0,\pi/4,\pi/2,3\pi/4\}} \mathcal{N}\Bigg(\bigoplus_{c}\bigoplus_{s=c+\delta}\mathcal{N}\Big(\mathbf{O}(c,s,\theta)\Big)\Bigg)
\end{align}
}

\textbf{Static saliency map.}  
The static saliency map is obtained by combining normalized conspicuity maps:
{\small
\begin{equation}
    \mathbf{SS} = \frac{1}{3} \Big( \mathcal{N}(\mathbf{\bar I}) + \mathcal{N}(\mathbf{\bar C}) + \mathcal{N}(\mathbf{\bar O}) \Big)
\end{equation}
}

\subsection{Motion saliency detection}



To extract motion saliency, we compute motion information using a biologically inspired approach based on the Hassenstein--Reichardt detector (Fig.~\ref{fig.1}c)~\cite{hassenstein_systemtheoretische_1956}. For each direction $D\in\{\text{left, right, up, down}\}$ and temporal offset $\tau\in\{1,3,7,15\}$, the motion response is defined as:
{\small
\begin{align}
\mathbf{M}(\sigma,t,D,\tau) &= \operatorname{ReLU}\Big(\exp\big(-\big|\mathbf{I_+}(\sigma,t)-\mathbf{I_+^\prime}(\sigma,t,D,\tau)\big|\big) \notag \\
& -\exp\big(-\big|\mathbf{I_+}(\sigma,t)-\mathbf{I_+^\prime}(\sigma,t,-D,\tau)\big|\big)\Big)
\end{align}
}
where $\mathbf{I_+^\prime}$ is $\mathbf{I_+}$ shifted by one pixel along $D$.

Motion conspicuity maps are computed using a center--surround operation across scales and directions:

{\small
\begin{align}
    \mathbf{\bar M}(\tau) = \frac{1}{2}\bigoplus_{c}\bigoplus_{s = c + \delta}\bigoplus_{D \in {\{\text{l,r,u,d}\}}} &\Big( \mathcal{N}\big(\mathbf{M_+}(c,s;\sigma,t,D,\tau)\big) \notag \\ 
    + &\mathcal{N}\big(\mathbf{M_-}(c,s;\sigma,t,D,\tau)\big) \Big)
\end{align}
}

The dynamic saliency map is then obtained by integrating normalized conspicuity maps across temporal scales, weighted by an exponential decay factor:
{\small
\begin{align}
    \mathbf{DS} &= \sum_{\tau\in{\{1,3,5,7\}}}\gamma^{\tau-1}\mathcal{N}\left(\mathbf{\bar M}(\tau)\right)
\end{align}
}

\subsection{Saliency fusion}

The final saliency map is computed as a second-order fusion of static and dynamic saliency maps (Fig.~\ref{fig.2}):
{\small
\begin{align}
&\mathbf{S} = a \cdot \mathcal{N}\big(\mathbf{SS}\cdot \mathbf{DS}\big) + b \cdot \mathcal{N}\big(\mathbf{SS}\big) + c \cdot \mathcal{N}\big(\mathbf{DS}\big)
\end{align}
}

where $(a,b,c)=(1,0.3,0.3)$ yields optimal performance (Table~\ref{tab:weights}).

{
\begin{table}[htb!]
\begin{center}
\caption{Quantitative comparison of different weights on DHF1K}
\vspace{-2mm} 
\label{tab:weights}
\resizebox{\linewidth}{!}
{
\begin{tabular}{c|c|c|c|c|c}
\hline
Weight(a,b,c)&CC&SIM&s-AUC&NSS&AUC-J\\
\hline
1.0, 0.2, 0.2 & 0.297 &0.171 &0.578 &1.57 &0.822\\
1.0, 0.3, 0.2& 0.300 & 0.193&0.578 &1.59 & 0.819 \\
1.0, 0.2, 0.3 & 0.301 & 0.178&0.576 &1.60 & 0.821 \\
1.0, 0.3, 0.3 & \textbf{0.307} &0.183 &\textbf{0.581} &\textbf{1.63} &\textbf{0.828}\\
1.0, 0.5, 0.3 & 0.299 & 0.190 &0.579 &1.58 & 0.818 \\
1.0, 0.3, 0.5 & 0.297 & 0.173& 0.574 & 1.60 & 0.820\\
1.0, 0.5, 0.5 & 0.301 &0.181 &0.578 &1.61 &0.821\\
1.0, 1.0, 1.0 & 0.300 &0.174 &0.578 &1.59 &0.821\\
1.0, 5.0, 5.0 & 0.298 &0.172 &0.577 &1.57 &0.821\\
0.0, 1.0, 0.0& 0.246 &0.147 &0.546 &1.66 &0.808\\
0.0, 0.0, 1.0& 0.282 &\textbf{0.198} &0.583 &1.53 &0.782\\
0.0, 0.5, 0.5 &0.297 &0.172 &0.577 &1.57 &0.821\\
\hline
\end{tabular}
}
\end{center}
\textbf{Bold} indicates the best performance.
\end{table}
}

\subsection{Fixation location prediction}

Fixation locations are derived from the master saliency map $\mathbf{S}$ in three steps. 

\begin{figure}[!ht]
    \centering
    {\includegraphics[width=\linewidth]{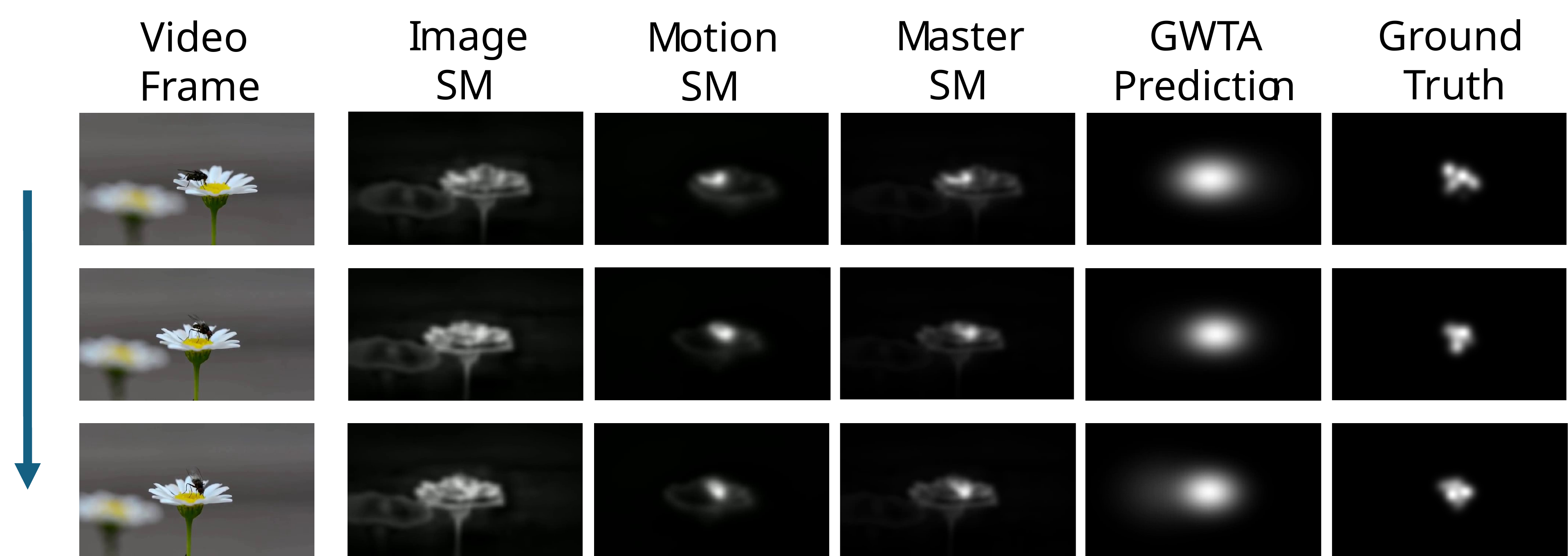}}
    \caption{
    Predicted saliency maps on an example video clip from the DHF1K dataset. From left to right: original frames, image-based saliency maps, motion-based saliency maps, combined master saliency maps, predicted fixations with GWTA, and human fixation ground truth. SM = saliency map, GWTA = Gaussian winner-take-all.
    }
    \label{fig.2}
    \vspace{-2mm}
\end{figure}

\textbf{(1) Gaussian-WTA (GWTA) modeling.}  
We extend the winner-take-all model with Gaussian-shaped foci of attention (Fig. \ref{fig.2}c). The Gaussian mask
{\small
\begin{align}
G(\boldsymbol{\mu}, \mathbf{\Sigma}) 
= \frac{1}{\sqrt{2\pi|\mathbf{\Sigma}|}}
\exp\!\left(-\tfrac{1}{2}(\boldsymbol{x}-\boldsymbol{\mu})^T 
\mathbf{\Sigma}^{-1}(\boldsymbol{x}-\boldsymbol{\mu})\right)
\end{align}
}
is optimized via gradient ascent to maximize 
{\small
\begin{align}
C(\boldsymbol{\mu}, \mathbf{\Sigma}) = \sum_{\boldsymbol{x}\in [H]\times [W]} S(\boldsymbol{x})\, G(\boldsymbol{x},\boldsymbol{\mu},\mathbf{\Sigma})
\end{align}
}


We iteratively update the $\boldsymbol{\mu}$ and $\Sigma$. To ensure numerical stability, $\mathbf{\Sigma}$ is set to be diagonal and regularized with $\lambda/\sigma$ ($\lambda=0.03\sqrt{W}$). 


\begin{equation}
    \boldsymbol{\mu}^{(n+1)} = \boldsymbol{\mu}^{(n)} + l_{\boldsymbol{\mu}}\frac{\partial C}{\partial\boldsymbol{\mu}}\Big|_{(\boldsymbol{\mu}^{(n)}, \boldsymbol{\Sigma}^{(n})}
\end{equation}

\begin{equation}
    \sigma_i^{(n+1)} = \sigma_i^{(n)} + l_{\sigma}\left(\frac{\partial C}{\partial \sigma_i}\Big|_{(\boldsymbol{\mu}^{(n)}, \boldsymbol{\Sigma}^{(n})} +\frac{\lambda}{\sigma_i}\right) 
\end{equation}

where $l_{\boldsymbol{\mu}}=0.1$ and $l_{\sigma}=4$ denote the update step lengths. Parameters are updated iteratively until reaching an empirical bound of 15 steps. The final fixation map is:

{\small
\begin{align}
\text{GWTA}(\boldsymbol{x}) = 
\sum_{i=1}^{N} S_i(\boldsymbol{x}) 
\exp\!\left(-\tfrac{1}{2}(\boldsymbol{x}-\hat{\boldsymbol{\mu}}_i)^T
\hat{\mathbf{\Sigma}}_i^{-1}(\boldsymbol{x}-\hat{\boldsymbol{\mu}}_i)\right)
\end{align}
}

where $N$ denotes the number of Gaussians and $\hat{\boldsymbol{\mu}}_i$ denotes the center of the $i$-th Gaussian. $N$ is increased iteratively until either the maximum of the residual saliency map falls below $0.2$ or $N$ reaches the upper limit of $12$ (matching the fixation count labeled by subjects in the DHF1K dataset).

\textbf{(2) Center prior.}  

To account for the center bias in the DHF1K dataset, a Gaussian center prior is applied:



{\small
\begin{align}
P(\boldsymbol{x}) = \exp\left[
-\frac{1}{2}\left(\boldsymbol{x} - \boldsymbol{\mu}\right)^\top
\begin{pmatrix}
    1/\sigma_x^2&0\\
    0&1/\sigma_y^2
\end{pmatrix}
\left(\boldsymbol{x} - \boldsymbol{\mu}\right)
\right]
\end{align}
}

where $\boldsymbol{\mu}=(W/2, H/2)$, $(\sigma_x, \sigma_y)=(W/3, H/3)$. The fixation map is updated as $\; F(\boldsymbol{x}) = \text{GWTA}(\boldsymbol{x}) \, P(\boldsymbol{x})$.

\textbf{(3) Temporal smoothing.}  
Temporal consistency is ensured using an exponentially weighted moving average (EWMA, Fig. \ref{fig.2}d):  
{\small
\begin{align}
\tilde{F}_t &= \alpha \, F_t + (1-\alpha)\,\tilde{F}_{t-1}, \quad \tilde{F}_0 = F_0
\end{align}
}
with $\alpha=0.9$.

\subsection{Human fixation clustering and labeling} 



To evaluate model performance, we analyzed human fixation patterns in DHF1K.
Fixations were sub-sampled (10\% of total), and 1,000 points per frame were clustered using DBSCAN ($\varepsilon=1.4$, $\text{min\_samples}=12$).
Each video was further categorized using YOLOv8~\cite{varghese_yolov8_2024} trained on OpenImages v7 into 13 object-based categories, enabling semantic-level performance analysis.


\section{Experiment Results}

\subsection{Experiment setup}

\textbf{Training/Testing Protocols.} We evaluate BIAS on the DHF1K dataset, a large, densely annotated benchmark for dynamic fixation prediction and video saliency~\cite{wang_revisiting_2018}. DHF1K contains 1,000 videos across seven categories (daily activity, sport, social activity, artistic performance, animal, artifact, landscape) with eye-tracking from 17 observers and standard evaluation splits and protocols. It also provides standardized evaluation protocols and maintains leaderboard performance using both heuristic- and learning-based models~\cite{wang_revisiting_2018}.

\textbf{Implementation.} BIAS is implemented in Python with a custom compiled dynamic library. The code is publicly available at \url{https://github.com/YatangLiLab/BIAS}. Runtime was measured on an AMD Ryzen 7 5800H using four CPU threads. 

\textbf{Baselines.} We compare BIAS to methods on the DHF1K leaderboard, covering both heuristic and learning-based video saliency approaches.

\textbf{Evaluation metrics.} Following DHF1K, we report five standard metrics: normalized scanpath saliency (NSS), similarity metric (SIM), Pearson's correlation coefficient (CC), area under the curve–Judd (AUC-J), and shuffled AUC (s-AUC). AUC values are reported based on 100 random permutations.


\subsection{Performance comparisons}

\begin{figure}[!ht]
\centering
\includegraphics[width=1\linewidth]{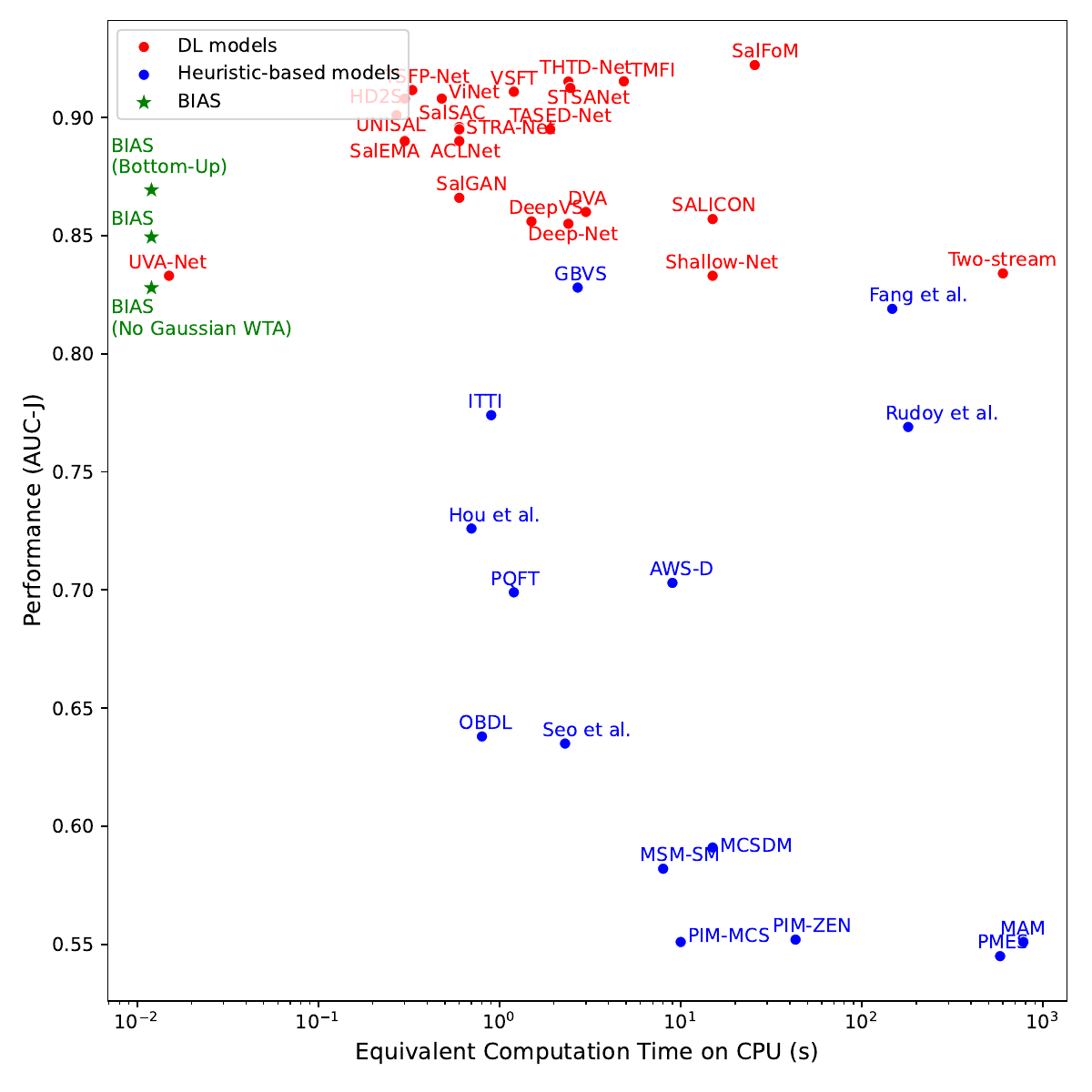}
\caption{Comparison of performance and runtime between BIAS and other models. 
}
\label{fig.3}
\vspace{-2mm} 
\end{figure}

We computed a master saliency map for each frame and applied GWTA to produce fixation predictions. Table~\ref{tab:performance} and Fig.~\ref{fig.3} summarize quantitative comparisons of performance and runtime.

BIAS outperforms all heuristic–based saliency models, with the GWTA further improving the performance. Notably, despite relying exclusively on bottom-up cues, it achieves performance comparable to roughly one-third of deep learning-based models that explicitly incorporate top-down attention, especially when evaluating video categories driven predominantly by bottom-up attention.

\begin{table}[!ht]
\caption{Quantitative comparison of different methods on DHF1K}
\vspace{-2mm} 
\resizebox{\linewidth}{!}
{
\begin{tabular}{c|c|c|c|c|c|c|c|c}
\hline
Method & AUC-J & SIM & s-AUC & CC & NSS & Size (MB) & DLM & Time$^\star$ (s) \\
\hline
SalFoM       &0.922&0.421&0.735&0.569&3.354&1574 &\checkmark &90 \\
TMFI         &0.915&0.407&0.731&0.546&3.146&234  &\checkmark &4.95 \\
THTD-Net     &0.915&0.406&0.730&0.548&3.139&220  &\checkmark &12 \\
STSANet      &0.912&0.383&0.723&0.529&3.010&643  &\checkmark &5.25 \\
TSFP-Net     &0.912&0.392&0.723&0.517&2.967&58.4 &\checkmark &1.65 \\
VSFT         &0.911&0.411&0.720&0.518&2.977&71.4 &\checkmark &6 \\
HD2S         &0.908&0.406&0.700&0.503&2.812&116  &\checkmark &4.5 \\
ViNet        &0.908&0.381&0.729&0.511&2.872&124  &\checkmark &2.4 \\
UNISAL       &0.901&0.390&0.691&0.490&2.776&15.5 &\checkmark &1.35 \\
SalSAC       &0.896&0.357&0.697&0.479&2.673&93.5 &\checkmark &3 \\
TASED-Net    &0.895&0.361&0.712&0.470&2.667&82   &\checkmark &9 \\
STRA-Net     &0.895&0.355&0.663&0.458&2.558&641  &\checkmark &3 \\
SalEMA       &0.890&0.466&0.667&0.449&2.574&364  &\checkmark &1.5 \\
ACLNet       &0.890&0.315&0.601&0.434&2.354&250  &\checkmark &3 \\
\makecell{\textbf{BIAS}\\\textbf{(Bottom-Up)}} & \textbf{0.869} & \textbf{0.237} & \textbf{0.577} & \textbf{0.358} & \textbf{1.851} & \textbf{0} & \textbf{×} & \textbf{0.012} \\
SalGAN       &0.866&0.262&0.709&0.370&2.043&130  &\checkmark &3 \\
DVA          &0.860&0.262&0.595&0.358&2.013&96   &\checkmark &15 \\
SALICON      &0.857&0.232&0.590&0.327&1.901&117  &\checkmark &75 \\
DeepVS       &0.856&0.256&0.583&0.344&1.911&344  &\checkmark &7.5 \\
Deep-Net     &0.855&0.201&0.592&0.331&1.775&103  &\checkmark &12 \\
\textbf{BIAS} & \textbf{0.849} & \textbf{0.221} & \textbf{0.561} & \textbf{0.323} & \textbf{1.670} & \textbf{0} & \textbf{×} & \textbf{0.012} \\
Two-stream   &0.834&0.197&0.581&0.325&1.632&315  &\checkmark &3000 \\
UVA-Net      &0.833&0.241&0.582&0.307&1.536&-    &\checkmark &0.058 \\
Shallow-Net  &0.833&0.182&0.529&0.295&1.509&2500 &\checkmark &15 \\
\makecell{\textbf{BIAS}\\\textbf{(No GWTA)}} & \textbf{0.828} & \textbf{0.183} & \textbf{0.582} & \textbf{0.307} & \textbf{1.626} & \textbf{0} & \textbf{×} & \textbf{0.011} \\
GBVS         &0.828&0.186&0.554&0.283&1.474& -     &× &2.7 \\
Fang et al.  &0.819&0.198&0.537&0.273&1.539& -     &× &147 \\
ITTI         &0.774&0.162&0.553&0.233&1.207& -     &× &0.9 \\
Rudoy et al. &0.769&0.214&0.501&0.285&1.498& -     &× &180 \\
Hou et al.   &0.726&0.167&0.545&0.150&0.847& -     &× &0.7 \\
AWS-D        &0.703&0.157&0.513&0.174&0.940& -     &× &9 \\
PQFT         &0.699&0.139&0.562&0.137&0.749& -     &× &1.2 \\
OBDL         &0.638&0.171&0.500&0.117&0.495& -     &× &0.8 \\
Seo et al.   &0.635&0.142&0.499&0.070&0.334& -     &× &2.3 \\
MCSDM        &0.591&0.110&0.500&0.047&0.247& -     &× &15 \\
MSM-SM       &0.582&0.143&0.500&0.058&0.245& -     &× &8 \\
PIM-ZEN      &0.552&0.095&0.498&0.062&0.280& -     &× &43 \\
PIM-MCS      &0.551&0.094&0.499&0.053&0.242& -     &× &10 \\
MAM          &0.551&0.108&0.500&0.041&0.214& -     &× &778 \\
PMES         &0.545&0.093&0.502&0.055&0.237& -     &× &579 \\
\hline
\end{tabular}
}

\vspace{1mm} 
$^\star$ Estimated runtime on an AMD Ryzen 7 5800H CPU. 

\textbf{Bold} highlights BIAS performance.
\label{tab:performance}
\end{table}

\subsection{Ablation study}

\begin{figure}[!ht]
    \centering
    \begin{minipage}[t]{0.45\linewidth}
        \centering
        \subfloat[]{\includegraphics[width=\linewidth]{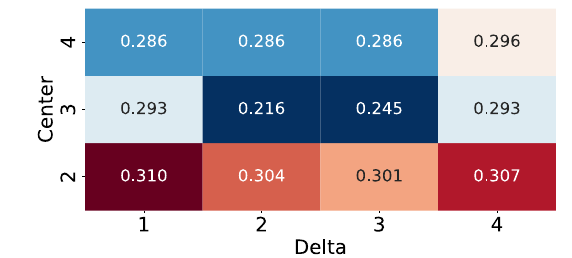}\label{fig.2b}} \\
        \vspace{2mm}
        \subfloat[]{\includegraphics[width=\linewidth]{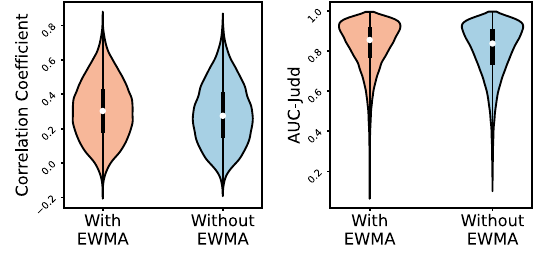}\label{fig.2d}}
    \end{minipage}%
    \hfill
    \begin{minipage}[t]{0.5\linewidth}
        \centering
        \subfloat[]{\includegraphics[width=\linewidth]{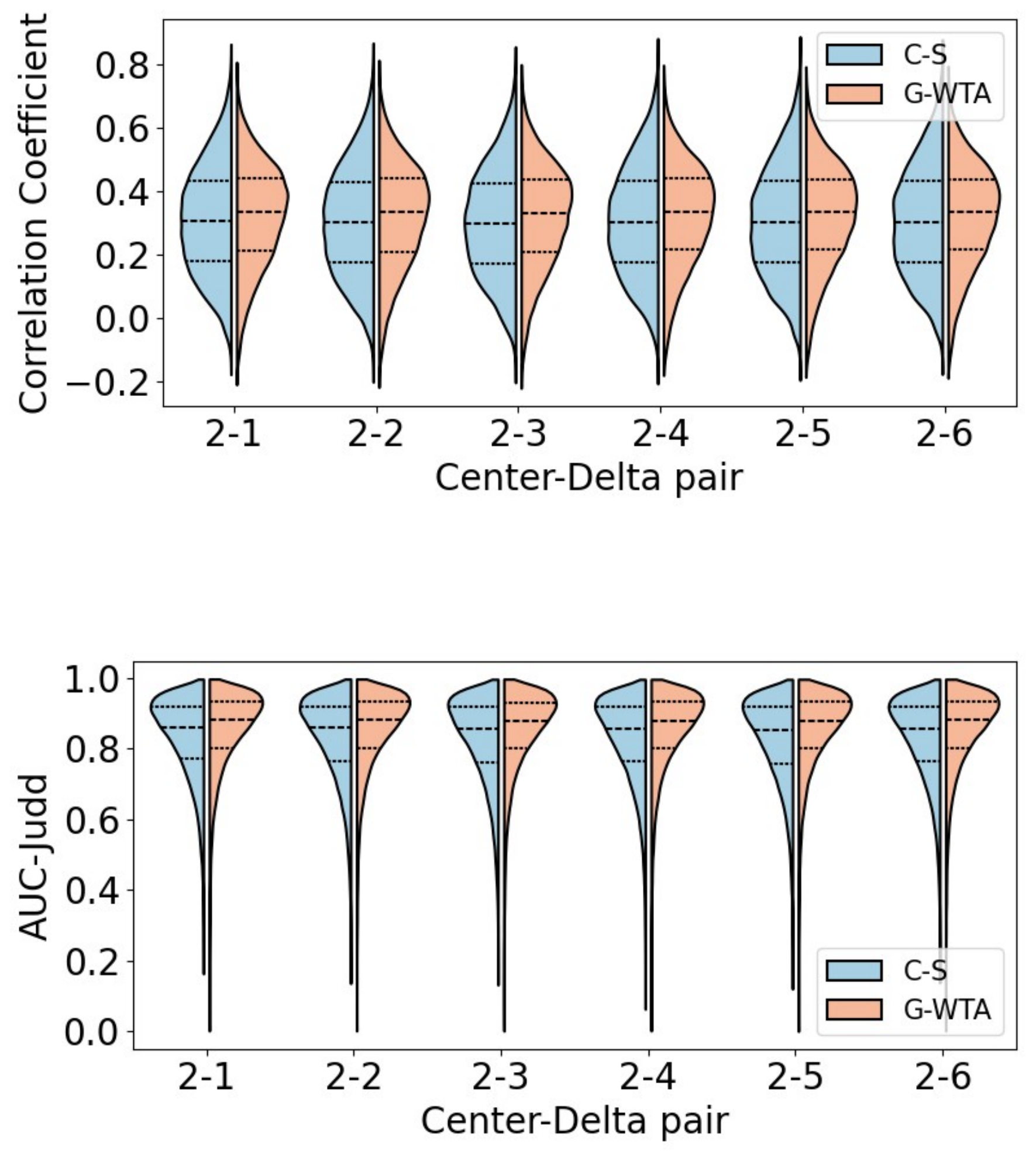}\label{fig.2c}}
    \end{minipage}
    \caption{
    (a) Correlation coefficients for different center--delta pairs.
    (b) Performance comparison with and without EWMA.
    (c) Performance comparison with and without GWTA across different center--delta pairs.
    }
    \label{fig:ablation}
    \vspace{-3mm}
\end{figure}

We ran ablations on the DHF1K validation set to quantify the contribution of key components: center–surround scale pairs, GWTA fixation selection, EWMA temporal smoothing, and the flicker motion cue (Table~\ref{tab:ablation} and Fig.~\ref{fig:ablation}).


\textbf{Center-delta pairs.} Prior work uses multiple center–surround combinations ($c\in\{2,3,4\}$, $\delta\in\{3,4\}$) to capture multi-scale contrast at increased cost~\cite{itti_model_1998}. On DHF1K, a single $(c,\delta)$ pair yields only a small performance drop; adding EWMA and GWTA largely eliminates this gap and can even surpass the multi-scale baseline, showing that accurate saliency can be extracted with reduced computation.


\textbf{GWTA (fixation selection).} GWTA consistently improves all metrics relative to alternatives such as L\'evy-flight sampling, a common way to model gaze shift\cite{Brockmann1999levy} \cite{Giuseppe2004Constrained} \cite{marlow2015temporal}, demonstrating better alignment with human fixations.


\textbf{EWMA (temporal smoothing).} Applying EWMA produces a clear and consistent performance improvement and reduces variability across $(c,\delta)$ settings, stabilizing frame-to-frame saliency predictions.


\textbf{Flicker cue.} Including a flicker-based motion feature slightly degrades performance in our pipeline, suggesting it is not complementary to the other motion channels we use.

\begin{table}[!htb]
\footnotesize
\begin{center}
\caption{Ablation study on DHF1K}
\vspace{-4mm} 
\label{tab:ablation}
\resizebox{\linewidth}{!}
{
\begin{tabular}{c|c|c|c|c|c}
\hline
Method                                     & AUC-J &SIM  &sAUC& CC & NSS\\
\hline
\makecell{BIAS (Bottom-Up)\\($c\in\{2\}$, $\delta\in\{1\}$)}                                 &\textbf{0.869}     &\textbf{0.237}     &0.577      &\textbf{0.358}     &\textbf{1.851}\\
\makecell{BIAS\\($c\in\{2\}$, $\delta\in\{4\}$)}               &0.849     &0.221     &0.561      &0.323     &1.670\\
\makecell{BIAS\\($c\in\{2\}$, $\delta\in\{1\}$)}               &0.849     &0.223     &0.560      &0.323     &1.670\\
\makecell{w/ L\'evy flight, w/o GWTA\\($c\in\{2\}$, $\delta\in\{4\}$)}     &0.846     &\textbf{0.237}     &0.559      &0.311     &1.626\\
\makecell{w/ EWMA, w/o GWTA\\($c\in\{2\}$, $\delta\in\{4\}$)}                 &0.835     &0.184     &\textbf{0.582}      &0.310     &1.632\\
\makecell{w/ EWMA, w/o GWTA\\($c\in\{2\}$, $\delta\in\{1\}$)}                 &0.828     &0.184     &\textbf{0.582}      &0.307     &1.625\\
\makecell{w/ EWMA, w/ GWTA\\($c\in\{2,3,4\}$, $\delta\in\{3,4\}$)}          &0.844     &0.226     &0.562      &0.319     &1.636       \\
\makecell{w/ EWMA, w/o GWTA\\($c\in\{2,3,4\}$, $\delta\in\{3,4\}$)}         &0.827     &0.184     &0.578      &0.304     &1.602\\
\makecell{w/o EWMA\&GWTA\\($c \in \{2\}$, $\delta \in\{4\}$)}      &0.802     &0.176     &0.571      &0.279     &1.495\\
\makecell{w/o EWMA\&GWTA\\($c \in \{2,3,4\}$, $\delta \in\{3,4\}$)}&0.818&0.189&0.578&0.299&1.580\\
\makecell{w/ flicker, w/o GWTA\\($c\in\{2\}$, $\delta\in\{4\}$)}      &0.828     &0.184     &0.572      &0.284     &1.470\\
\hline
\end{tabular}
}
\end{center}
\textbf{Bold} indicates the best performance.
\end{table}

\subsection{Computational cost}


The DHF1K benchmark reports GPU runtimes (Titan X), while BIAS runs on CPU. We empirically derived a GPU-to-CPU conversion factor of 31.7× by benchmarking several public models and used this factor for comparison. Our approach achieves the fastest runtime, demonstrating its suitability for real-time CPU deployment.


\subsection{Influence of video contents on performance}


Because BIAS focuses exclusively on bottom-up saliency, we hypothesize that it performs better on clips dominated by stimulus-driven cues. To test this, we analyzed per-video variability and correlated performance with semantic categories (YOLO-v8 \cite{varghese_yolov8_2024} trained on OpenImages v7 dataset \cite{OpenImages}) and fixation cluster counts (DBSCAN \cite{DBSCAN_1996} on sub-sampled fixations; $\varepsilon=1.4$).

\begin{figure}
\centering
    \subfloat[]{\includegraphics[width=1\linewidth]{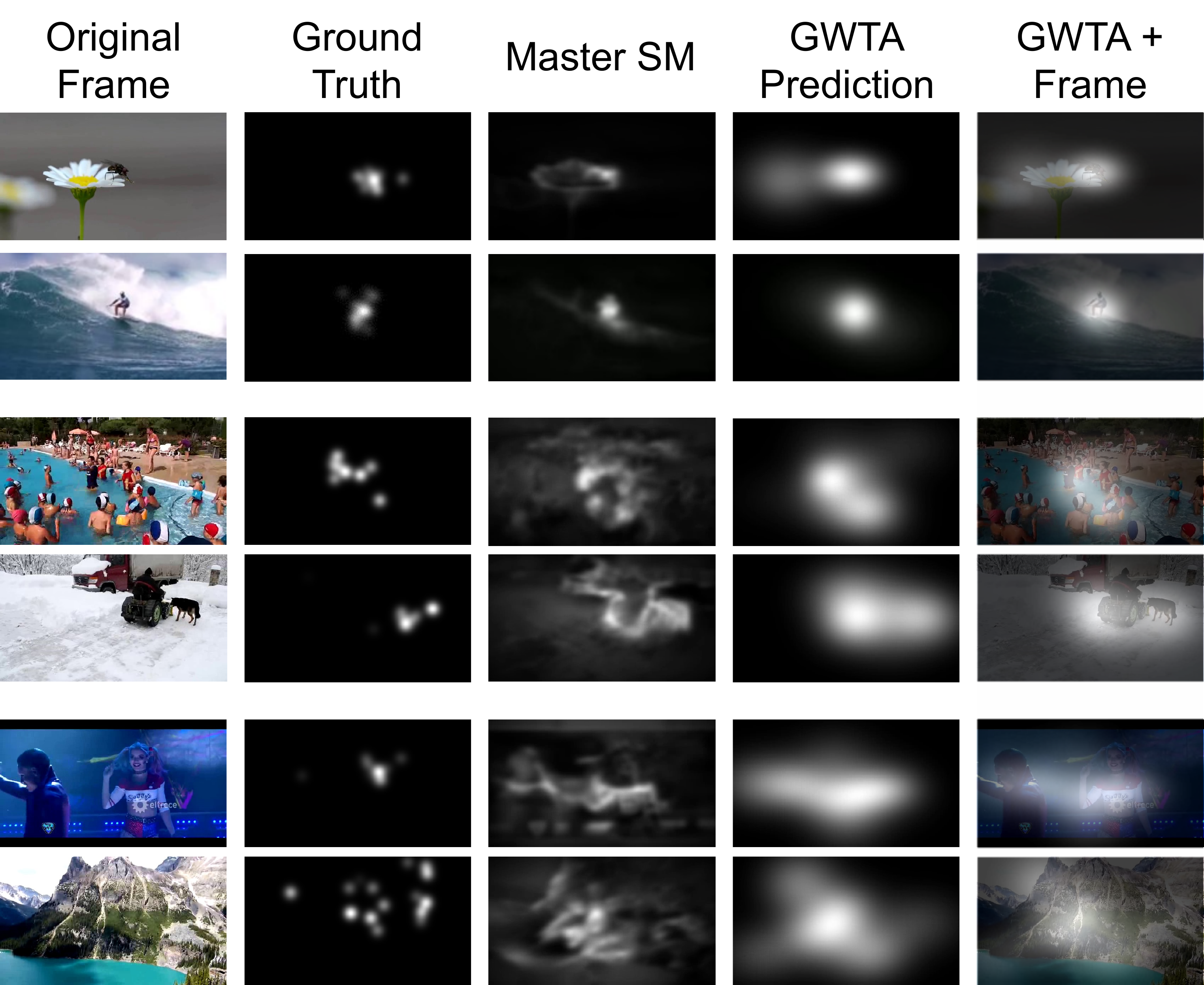}}
    \newline
    \subfloat[]{\includegraphics[width=0.75\linewidth]{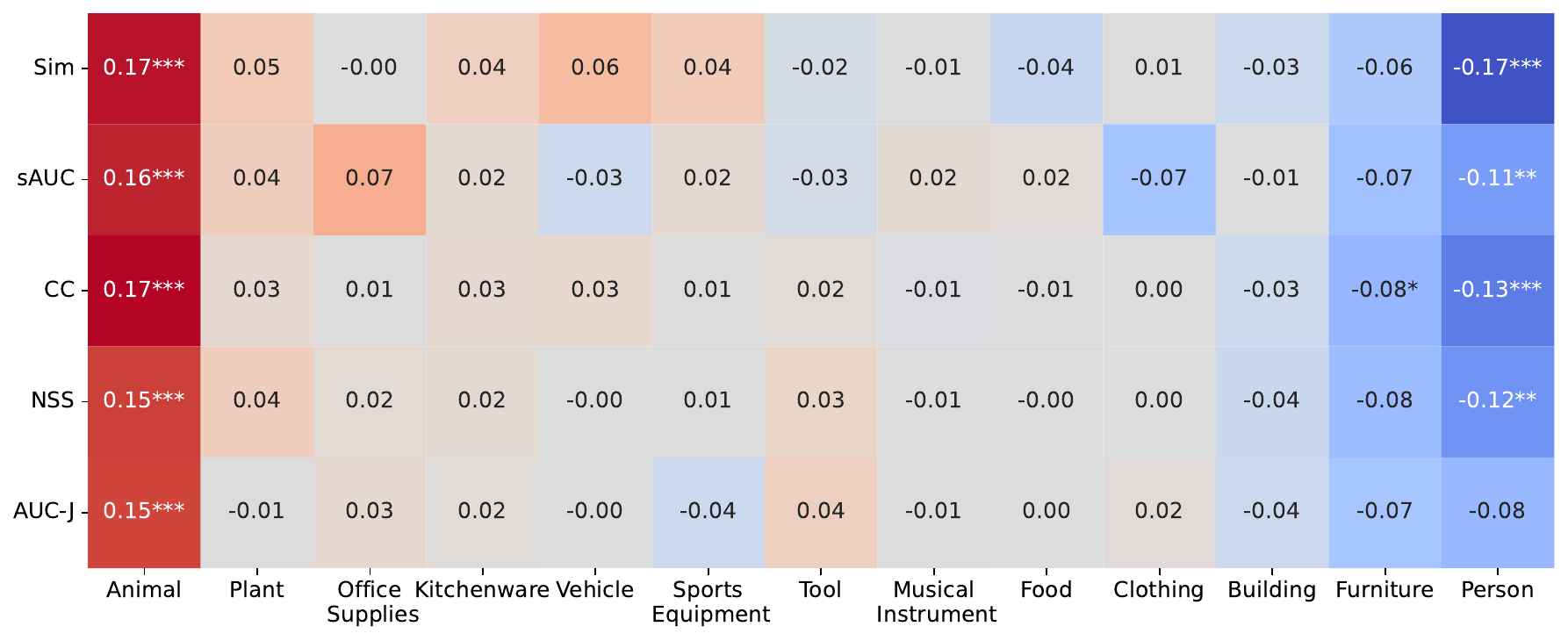}}
    \subfloat[]{\includegraphics[width=0.215\linewidth]{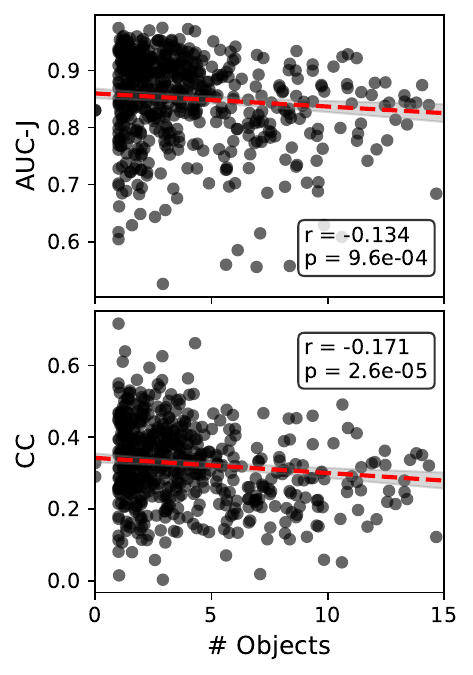}}
    \newline
    \subfloat[]{\includegraphics[width=1\linewidth]{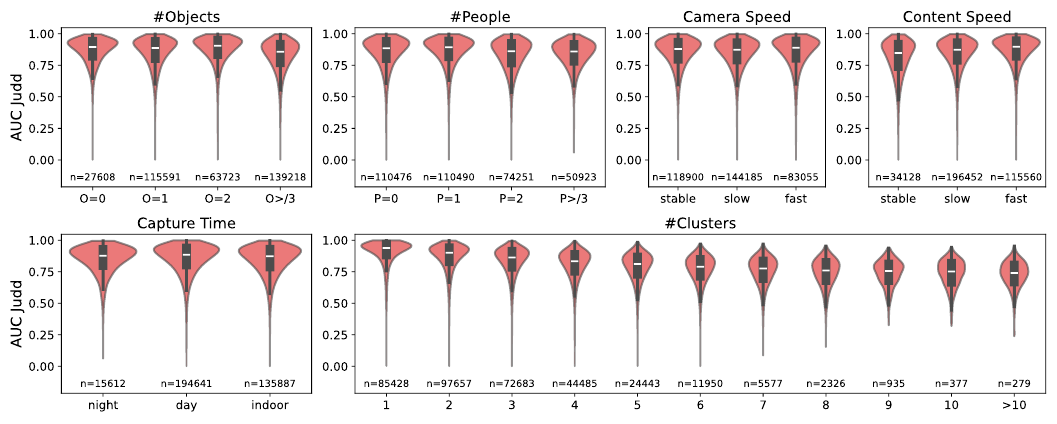}}
    \newline
    \subfloat[]{\includegraphics[width=1\linewidth]{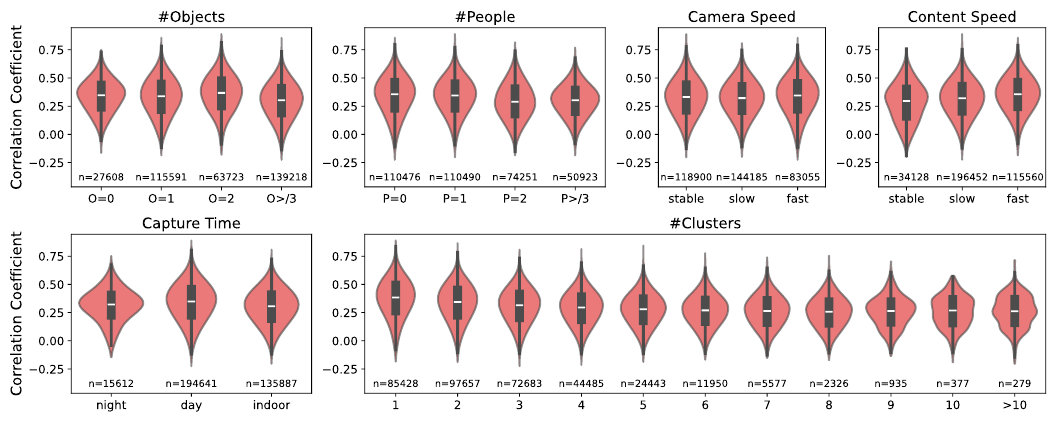}}
    \newline
    \caption{
    (a) From left to right: original frames, human fixation ground truth, master saliency maps, predicted fixations with GWTA, and GWTA predictions overlaid on the original frames. The top two rows show examples of well-predicted saliency maps; the middle two rows show examples of moderate performance; the bottom two rows show examples of low performance.
    (b) Performance metrics for videos with distinct COCO-labeled contents. *** $p< 0.001$, ** $p< 0.01$, * $p< 0.05$.
    (c) Correlation of mean performance metrics with the average numbers of  YOLO-detected objects for each video. $r$ is the Pearson correlation coefficient.
    (d--e) Performance metrics of predicted saliency maps in different video categories. 
    The number of objects and people, camera speed, content speed, and capture time were obtained from annotation data of the DHF1K dataset. The number of clusters was obtained using the DBSCAN clustering algorithm.
    }
\label{fig.5}
\end{figure}

Our hypothesis is supported by several findings. First, videos containing animals show higher performance, likely because a small number of moving animals generate strong bottom-up cues (Fig.~\ref{fig.5}a). Second, performance decreases as the number of objects or people increases (Fig.~\ref{fig.5}c--e). This trend is consistent whether quantified by detected object counts or by fixation clusters, and aligns with known cognitive limits on attentional tracking (e.g., up to five objects). Finally, higher object motion speeds tend to improve performance by strengthening motion cues, whereas slow global (camera-induced) motion can reduce the signal-to-noise ratio in motion channels (Fig.~\ref{fig.5}d--e).


\subsection{Traffic accident anticipation}
Given BIAS's low latency and its inherent sensitivity to sudden, stimulus-driven visual anomalies, we further evaluated its real-world utility in a highly time-critical downstream task: traffic accident anticipation (TAA), using the Traffic Accident Benchmark for Causality Recognition~\cite{you_traffic_2020}. This benchmark provides a standardized dataset for analyzing accident causality and includes Kinetics-I3D features~\cite{carreira_quo_2017} as inputs for analysis algorithms.

\begin{figure}[!ht]
\centering
{\includegraphics[width=1\linewidth]{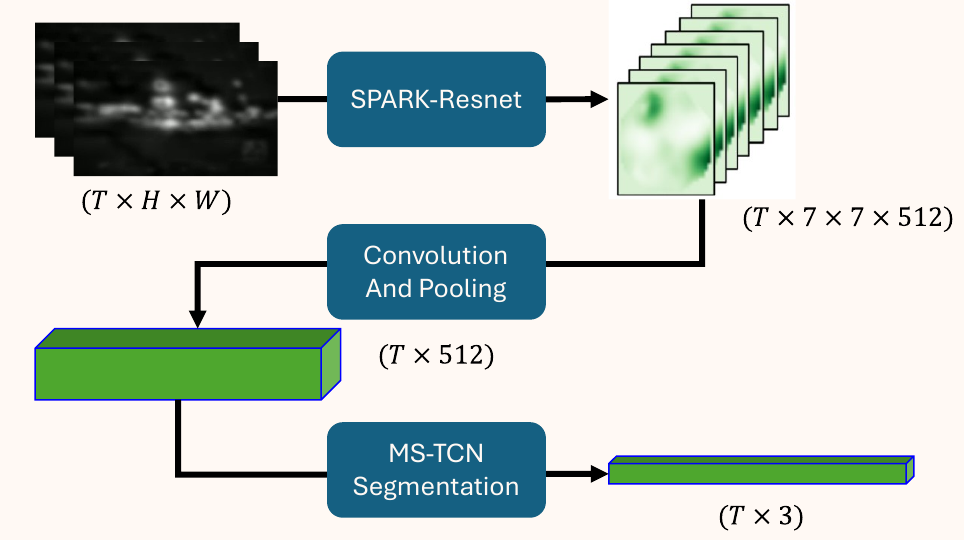}}
\caption{The input is a sequence of predicted saliency maps. SPARK-ResNet extracts only spatial features, which are then processed via convolution and pooling before being fed into the MS-TCN for temporal segmentation.}
\label{fig.traffic_model}
\end{figure}

Instead of relying on these supervised Kinetics-I3D features, we adopted a self-supervised approach (Fig. \ref{fig.traffic_model}). Specifically, we first generated saliency maps from the original traffic video using BIAS. These maps were then compressed from high-dimensional image representations into lower-dimensional,  semantically enriched features via the SparK framework~\cite{tian_designing_2023}. The resulting features were passed through a convolutional bottleneck module for dimensionality reduction, followed by an MS-TCN module~\cite{farha_ms-tcn_2019} for temporal reasoning and accident anticipation. This architecture preserves discriminative information while mitigating overfitting.

\textbf{Bottleneck design.} We tested several bottleneck modules (AvgPool, Conv, Conv+AvgPool). The hybrid Conv+AvgPool design achieves the best balance between representational capacity and robustness (Table~\ref{tab:design}). Tuning the hidden dimensionality further reveals that moderate channel sizes (e.g., 128) yield optimal performance: smaller dimensions tend to underfit, whereas larger ones increase the risk of overfitting (Table~\ref{tab:dimension}).




\textbf{Accident causality recognition.} To evaluate the effectiveness of BIAS in accident causality recognition, we compared Intersection-over-Union (IoU) scores for cause and effect video segmentation using BIAS-SparK features and Kinetics-I3D (RGB) features (Table~\ref{tab:iou_results}). BIAS-SparK achieves substantially higher performance in effect segmentation, consistently yielding improved IoU across multiple thresholds relative to RGB features. In contrast, its performance in cause segmentation is lower than that of the Kinetics-I3D baseline. 

Replacing BIAS with either the original Itti model or the state-of-the-art deep-learning model SalFoM\cite{moradi_salfom_2025} results in degraded performance in both cause and effect segmentation. These results suggest that bottom-up saliency is critical for distinguishing causal structure in traffic accidents, whereas the inclusion of top-down components may compromise this capability.

Notably, the lower cause IoU observed for BIAS-SparK does not necessarily reflect weaker causal inference. Instead, we hypothesize this result may be attributed to a temporal misalignment: BIAS-SparK's predictions often precede the annotated onset of causal events, thereby reducing overlap with the ground truth.

\begin{table}[ht!]
\centering
\caption{Performance on traffic accident causality recognition
}
\vspace{-1mm}
\footnotesize
\resizebox{\linewidth}{!}
{
\begin{tabular}{c|c|c|c|c}
\hline
Model & {Kinetics-I3D} & {BIAS-SparK}& {Itti-SparK}\cite{itti_model_1998}& {SalFoM-Spark}\cite{moradi_salfom_2025} \\
\hline
Cause IoU $\ge0.1$ & \textbf{0.538} & 0.513 & 0.172&0.455      \\
Cause IoU $\ge0.3$ & \textbf{0.415} & 0.323 & 0.068&0.293      \\
Cause IoU $\ge0.5$ & \textbf{0.201} & 0.143 & 0.007&0.111      \\
Cause IoU $\ge0.7$ & \textbf{0.061} & 0.054 & 0.003&0.021      \\
\hline
Effect IoU $\ge 0.1$ & 0.638 & \textbf{0.796} & 0.462&0.648\\
Effect IoU $\ge 0.3$ & 0.462 & \textbf{0.606} & 0.293&0.483\\
Effect IoU $\ge 0.5$ & 0.276 & \textbf{0.348} & 0.139&0.243\\
Effect IoU $\ge 0.7$ & 0.125 & \textbf{0.136} & 0.046&0.075\\
\hline
\end{tabular}
}

\begin{tablenotes}
    \footnotesize
    \item \textbf{bold}: best performance.
\end{tablenotes}
\label{tab:iou_results}
\end{table}

\begin{figure}[t]
\centering
\subfloat[]{\includegraphics[width=1\linewidth]{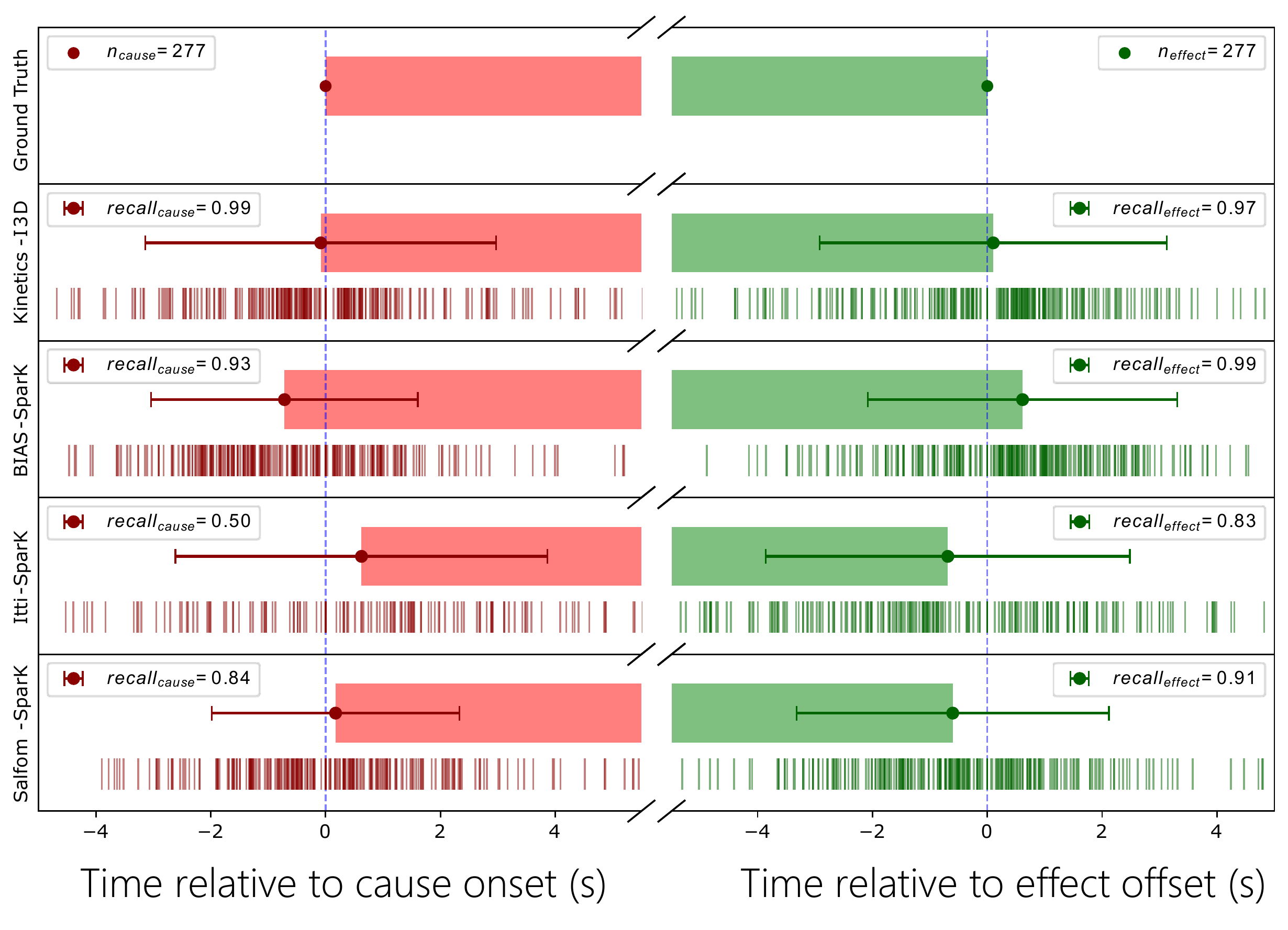}}
    \newline
\subfloat[]{\includegraphics[width=1\linewidth]{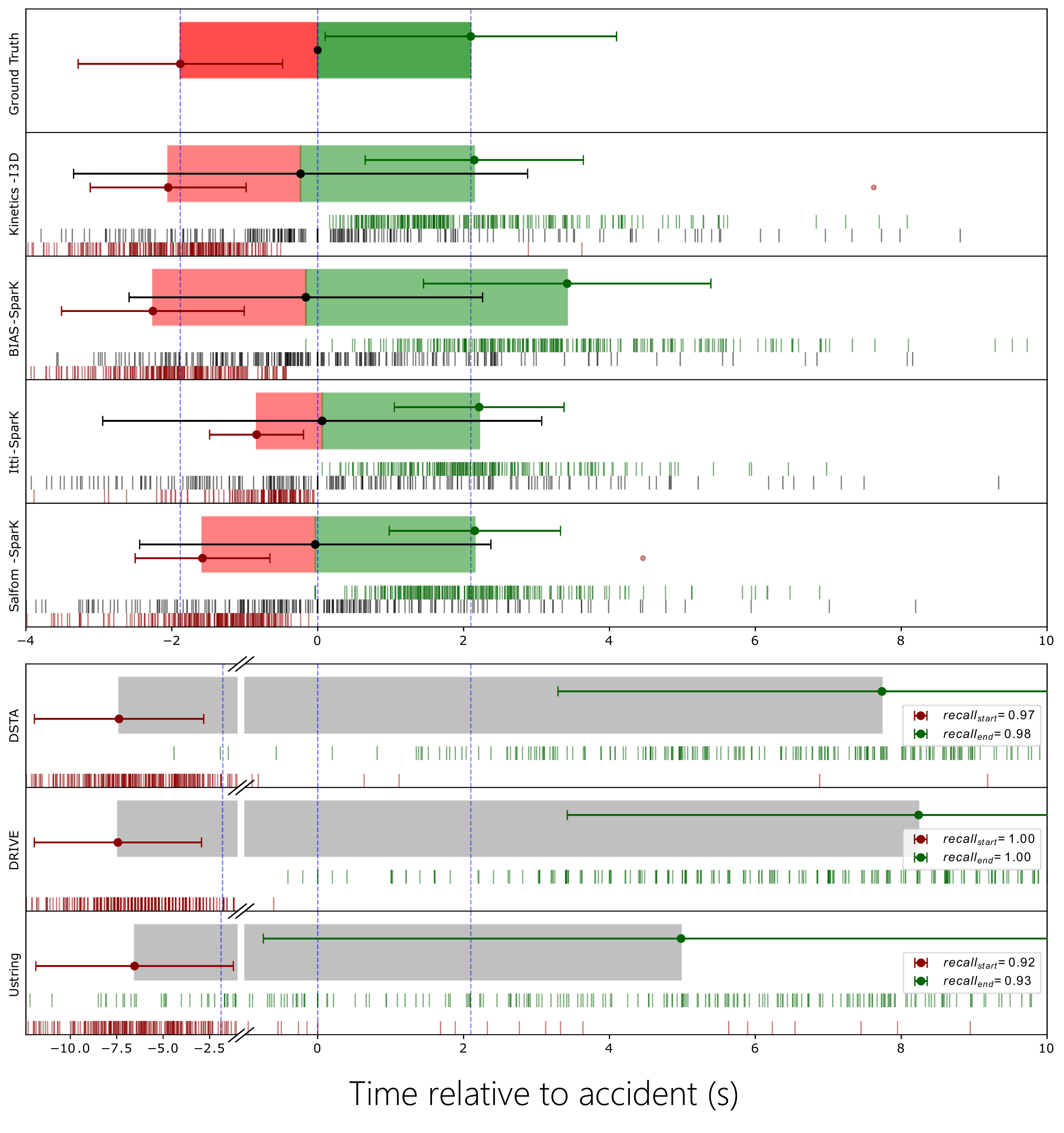}}

\caption{Comparison of predicted times across different models. (a) Predicted time relative to cause onset and effect offset. For cause onset, Kinetics-I3D: $-0.08 \pm 3.05$, $p = 0.65$ (one-sample $t$-test against 0); BIAS-SparK: $-0.72 \pm 2.32$, $p = 1.1 \times 10^{-6}$; Itti-SparK: $0.63 \pm 3.24$; SalFoM-SparK: $0.17 \pm 2.16$. For effect offset, Kinetics-I3D: $0.10 \pm 3.02$; BIAS-SparK: $0.61 \pm 2.69$; Itti-SparK: $-0.69 \pm 3.17$; SalFoM-SparK: $-0.60 \pm 2.72$. (b) Predicted TTAs by different methods. }
\label{fig.traffic_pred}
\vspace{-3mm}
\end{figure}

\textbf{Lead time analysis.} To test this hypothesis and evaluate BIAS's performance in accident anticipation, we compared the predicted cause onset with the labeled ground truth and defined lead time (LT) as their difference $t_{\text{lead}} = t_{\text{label}} - t_{\text{predict}}$. Supporting our hypothesis, BIAS-SparK features yield significantly earlier predictions, whereas predictions using Kinetics-I3D features are close to the ground truth (Fig.~\ref{fig.traffic_pred}(a) and Table~\ref{tab: SOTATAA}). This suggests that bottom-up saliency can reveal pre-incident cues before semantic labels of cause onset. In contrast, Itti-Spark and SalFoM-SparK models show later LT than the Kinetics-I3D model. Notably, these two models also detect fewer causes and effects, as indicated by their lower recall scores. Additionally, BIAS-SparK also shows a later predicted effect offset than other models.


To further assess predictive performance, we also evaluated BIAS using time to accident (TTA), a commonly used metric in traffic accident anticipation models. We compared BIAS-SparK with several state-of-the-art models (Fig.~\ref{fig.traffic_pred}(b) and Table~\ref{tab: SOTATAA}), including DRIVE~\cite{BaoICCV2021DRIVE}, UString~\cite{bao_uncertainty-based_2020}, and DSTA~\cite{karim_dynamic_2022}. 
Although these methods report longer TTA, their predictive reliability is substantially lower, as evidenced by the small fraction of predictions achieving high IoU. In fact, their accuracy barely exceeds random chance, as demonstrated by a randomization procedure that preserves per-video accident frame counts while randomly shuffling prediction labels.

Overall, BIAS-SparK achieves a favorable balance between early anticipation and prediction accuracy, delivering strong performance in both causality recognition and traffic accident anticipation.

\begin{table}[t!]
\centering
\caption{Traffic Accident Prediction IoU and Anticipation}
\vspace{-1mm}
\footnotesize
\resizebox{\linewidth}{!}
{
\begin{tabular}{c|c|c|c|c|c|c}
\hline
Model            & IoU $\ge$ 0.1 & IoU $\ge$ 0.3 & IoU $\ge$ 0.5 & IoU $\ge$ 0.7 & mTTA (s)  & mLT (s) \\
\hline
BIAS-SparK            & \textbf{0.89}      & \textcolor{red}{\textbf{0.78}}      & \textcolor{red}{\textbf{0.57}}      & 0.24      & \textbf{2.26}  & \textbf{0.72}  \\
Kinetics-I3D             & 0.83      & 0.73      & 0.53      & \textcolor{red}{\textbf{0.26}}      & 2.05  & 0.08  \\
Itti-SparK             & 0.58      & 0.42      & 0.19      & 0.06      & 0.83  & -0.62  \\
SalFoM-SparK           & 0.77      & 0.68      & 0.44      & 0.16      & 1.58  & -0.17  \\
\hline
DRIVE$^\star$ & \textcolor{red}{0.93}      & 0.30      & 0.05      & 0.02      & {8.01}  & {6.73}  \\
rand DRIVE$^\dagger$  & \textcolor{red}{0.93}      & 0.37      & 0.11      & 0.03      & {7.04}  & {5.73}  \\
Ustring$^\star$& 0.52      & 0.14      & 0.02      & 0.00      & {6.89}  & {5.24}  \\
rand Ustring$^\dagger$& 0.50      & 0.28      & 0.10      & 0.03      & {3.90}  & {1.60}  \\
DSTA$^\star$   & \textcolor{red}{0.93}      & 0.31      & 0.06      & 0.01      & {7.98}  & {6.58}  \\
rand DSTA$^\dagger$   & \textcolor{red}{0.93}      & 0.31      & 0.06      & 0.01      & \textcolor{red}{8.22}  & \textcolor{red}{6.86}  \\
\hline
\end{tabular}
}

\begin{tablenotes}
    \footnotesize
    \item \textcolor{red}{Red}: best overall; \textbf{bold}: best among MST-CN methods.
    \item[$\star$] Accident probability threshold = 0.5.
    \item[$\dagger$] Mean over 10 random shuffles.
\end{tablenotes}

\label{tab: SOTATAA}
\vspace{-2mm}
\end{table}

\section{Discussion and Conclusion}

In this work, we present \textbf{BIAS}, a fast and biologically inspired model for video saliency detection. By combining a retina-inspired motion detector with a Gaussian Winner-Take-All fixation selector, BIAS generates interpretable spatiotemporal saliency maps with millisecond-scale latency. BIAS achieves orders of magnitude greater computational efficiency than deep networks while outperforming classic heuristic methods and approaching the performance of modern learning-based models on the DHF1K benchmark.

When applied to traffic accident analysis, BIAS captures pre-accident visual cues and enables significantly earlier detection of accident causes, highlighting the critical role of bottom-up attention in safety-critical prediction and supporting the idea that low-level saliency signals can precede semantically defined causal events.


Limitations include reduced performance on tasks that require high-level semantic reasoning and potential sensitivity to complex object interactions or cluttered scenes. Future work will explore integrating top-down attention and task-driven modules, as well as deploying BIAS in real-world resource-constrained applications such as robotics and autonomous driving.

Overall, BIAS demonstrates that biologically motivated mechanisms remain a competitive, interpretable, and efficient strategy for real-time applications.

\newpage
{\appendix[Studies on Hyper-parameters Selection]

\setcounter{table}{0}   
\setcounter{figure}{0}
\setcounter{section}{0}
\setcounter{equation}{0}
\renewcommand{\thetable}{A\arabic{table}}
\renewcommand{\thefigure}{A\arabic{figure}}
\renewcommand{\thesection}{A\arabic{section}}
\renewcommand{\theequation}{A\arabic{equation}}

\begin{table}[!htb]
    \centering
    \caption{Ablation study on bottleneck module design. All hidden dimensions = 128.}
    \footnotesize
    \begin{tabular}{c|c|c|c}
    \hline
    \thead{Module} & \thead{2-Layer Conv} & \thead{Conv + AvgPool} & \thead{AvgPool Only} \\
    \hline
    Cause IoU $>0.1$ & 0.391 & \textbf{0.513} & 0.204 \\
    Cause IoU $>0.3$ & 0.237 & \textbf{0.323} & 0.086 \\
    Cause IoU $>0.5$ & 0.104 & \textbf{0.143} & 0.028 \\
    Cause IoU $>0.7$ & 0.025 & \textbf{0.054} & 0.007 \\
    \hline
    Effect IoU $>0.1$ & 0.556 & \textbf{0.796} & 0.243 \\
    Effect IoU $>0.3$ & 0.394 & \textbf{0.606} & 0.129 \\
    Effect IoU $>0.5$ & 0.190 & \textbf{0.348} & 0.046 \\
    Effect IoU $>0.7$ & 0.079 & \textbf{0.136} & 0.021 \\
    \hline
    \end{tabular}
    \label{tab:design}
\end{table}

\begin{table}[!htb]
    \centering
    \caption{Effect of hidden dimension size in the 1-layer convolutional bottleneck.}
    \footnotesize
    \begin{tabular}{c|c|c|c|c}
    \hline
    Hidden Dim & 32 & 64 & 128 & 256 \\
    \hline
    Cause IoU $>0.1$ & 0.502 & 0.455 & \textbf{0.513} & 0.455 \\
    Cause IoU $>0.3$ & \textbf{0.333} & 0.304 & 0.323 & 0.287 \\
    Cause IoU $>0.5$ & 0.136 & 0.125 & 0.143 & \textbf{0.147} \\
    Cause IoU $>0.7$ & 0.035 & 0.025 & \textbf{0.054} & 0.039 \\
    \hline
    Effect IoU $>0.1$ & 0.724 & 0.713 & \textbf{0.796} & 0.742 \\
    Effect IoU $>0.3$ & 0.584 & 0.559 & \textbf{0.606} & 0.563 \\
    Effect IoU $>0.5$ & \textbf{0.355} & 0.351 & 0.348 & 0.341 \\
    Effect IoU $>0.7$ & 0.104 & 0.096 & \textbf{0.136} & 0.097 \\
    \hline
    \end{tabular}
    \label{tab:dimension}
\end{table}
}

\newpage
\bibliographystyle{IEEEtran}
\bibliography{main_v5}

@String(PAMI = {IEEE Trans. Pattern Anal. Mach. Intell.})

@String(IJCV = {Int. J. Comput. Vis.})

@String(CVPR= {IEEE Conf. Comput. Vis. Pattern Recog.})

@String(ICCV= {Int. Conf. Comput. Vis.})

@String(ECCV= {Eur. Conf. Comput. Vis.})

@String(NIPS= {Adv. Neural Inform. Process. Syst.})

@String(TIP  = {IEEE Trans. Image Process.})

@String(TMM  = {IEEE Trans. Multimedia})

@String(ACMMM= {ACM Int. Conf. Multimedia})

@String(ICASSP=	{ICASSP})

@String(ICIP = {IEEE Int. Conf. Image Process.})

@String(ACCV  = {ACCV})

@String(AAAI = {AAAI})

@String(SPL	= {IEEE Sign. Process. Letters})

@String(VR   = {Vis. Res.})

@String(JOV	 = {J. Vis.})

@String(PAMI  = {IEEE TPAMI})

@String(IJCV  = {IJCV})

@String(CVPR  = {CVPR})

@String(ICCV  = {ICCV})

@String(ECCV  = {ECCV})

@String(NIPS  = {NeurIPS})

@String(TIP   = {IEEE TIP})

@String(TCSVT = {IEEE TCSVT})

@String(TMM   =	{IEEE TMM})

@String(ACMMM = {ACM MM})

@String(ICIP  = {ICIP})

@article{petersen_attention_2012,
	title = {The Attention System of the Human Brain: 20 Years After},
	volume = {35},
	issn = {0147-006X},
	shorttitle = {The Attention System of the Human Brain},
	url = {https://www.annualreviews.org/doi/10.1146/annurev-neuro-062111-150525},
	doi = {10.1146/annurev-neuro-062111-150525},
	number = {1},
	urldate = {2020-09-27},
	journal = {{Annu.} {Rev.} of {Neurosci.}},
	author = {Petersen, Steven E. and Posner, Michael I.},
	month = jun,
	year = {2012},
	Mynote = {Publisher: Annual Reviews},
	pages = {73--89},
}

@article{koch_how_2006,
	title = {How Much the Eye Tells the Brain},
	volume = {16},
	issn = {0960-9822},
	url = {https://www.cell.com/current-biology/abstract/S0960-9822(06)01639-3},
	doi = {10.1016/j.cub.2006.05.056},
	language = {English},
	number = {14},
	urldate = {2020-08-01},
	journal = {{Curr.} {Biol.}},
	author = {Koch, Kristin and McLean, Judith and Segev, Ronen and Freed, Michael A. and Berry, Michael J. and Balasubramanian, Vijay and Sterling, Peter},
	month = jul,
	year = {2006},
	Mynote = {Publisher: Elsevier},
	pages = {1428--1434},
}

@article{borji_state---art_2013,
	title = {State-of-the-Art in Visual Attention Modeling},
	volume = {35},
	issn = {1939-3539},
	doi = {10.1109/TPAMI.2012.89},
	number = {1},
	journal = PAMI,
	author = {Borji, Ali and Itti, Laurent},
	month = jan,
	year = {2013},
	Mynote = {Conf.Name: IEEE IEEE TPAMI},
	pages = {185--207},
}

@article{kelly_information_1962,
	title = {Information Capacity of a Single Retinal Channel},
	volume = {8},
	issn = {0096-1000},
	doi = {10.1109/TIT.1962.1057716},
	number = {3},
	journal = {IRE Trans. Inf. Theory},
	author = {Kelly, D.},
	month = apr,
	year = {1962},
	pages = {221--226},
}

@article{koch_shifts_1985,
	title = {Shifts in Selective Visual Attention: Towards the Underlying Neural Circuitry},
	volume = {4},
	issn = {0721-9075},
	shorttitle = {Shifts in Selective Visual Attention},
	language = {eng},
	number = {4},
	journal = {Human Neurobiology},
	author = {Koch, C. and Ullman, S.},
	year = {1985},
	pages = {219--227},
}

@article{itti_computational_2001,
	title = {Computational Modelling of Visual Attention},
	volume = {2},
	copyright = {2001 Nature Publishing Group},
	issn = {1471-0048},
	url = {https://www.nature.com/articles/35058500},
	doi = {10.1038/35058500},
	language = {en},
	number = {3},
	urldate = {2020-07-27},
	journal = {Nat. Rev. Neurosci.},
	author = {Itti, Laurent and Koch, Christof},
	month = mar,
	year = {2001},
	Mynote = {Number: 3 Publisher: Nature Publishing Group},
	pages = {194--203},
}

@article{itti_model_1998,
	title = {A Model of Saliency-Based Visual Attention for Rapid Scene Analysis},
	volume = {20},
	issn = {1939-3539},
	doi = {10.1109/34.730558},
	number = {11},
	journal = PAMI,
	author = {Itti, L. and Koch, C. and Niebur, E.},
	year = {1998},
	pages = {1254--1259},
}

@article{zhaoping_new_2019,
	series = {Computational {Neuroscience}},
	title = {A New Framework for Understanding Vision From the Perspective of the Primary Visual Cortex},
	volume = {58},
	issn = {0959-4388},
	url = {https://www.sciencedirect.com/science/article/pii/S0959438819300042},
	doi = {10.1016/j.conb.2019.06.001},
	language = {en},
	urldate = {2022-01-18},
	journal = {{Curr.} {Opin.} in {Neurobio.}},
	author = {Zhaoping, Li},
	year = {2019},
	pages = {1--10},
}

@article{treisman_feature-integration_1980,
	title = {A Feature-Integration Theory of Attention},
	volume = {12},
	issn = {0010-0285},
	url = {http://www.sciencedirect.com/science/article/pii/0010028580900055},
	doi = {10.1016/0010-0285(80)90005-5},
	language = {en},
	number = {1},
	urldate = {2020-11-06},
	journal = {Cogn. Psychol.},
	author = {Treisman, Anne M. and Gelade, Garry},
	year = {1980},
	pages = {97--136},
}

@article{wang_revisiting_2021,
	title = {Revisiting Video Saliency Prediction in the Deep Learning Era},
	volume = {43},
	issn = {1939-3539},
	url = {https://ieeexplore.ieee.org/document/8744328},
	doi = {10.1109/TPAMI.2019.2924417},
	number = {1},
	urldate = {2024-10-16},
	journal = PAMI,
	author = {Wang, Wenguan and Shen, Jianbing and Xie, Jianwen and Cheng, Ming-Ming and Ling, Haibin and Borji, Ali},
	month = jan,
	year = {2021},
	Mynote = {Conf.Name: IEEE IEEE TPAMI},
	pages = {220--237},
}

@article{zheng_unbearable_2025,
	title = {The Unbearable Slowness of Being: Why Do We Live at 10 Bits/s?},
	volume = {113},
	issn = {0896-6273},
	shorttitle = {The Unbearable Slowness of Being},
	url = {https://www.cell.com/neuron/abstract/S0896-6273(24)00808-0},
	doi = {10.1016/j.neuron.2024.11.008},
	language = {English},
	number = {2},
	urldate = {2025-02-17},
	journal = {Neuron},
	author = {Zheng, Jieyu and Meister, Markus},
	year = {2025},
	pages = {192--204},
}

@inproceedings{moradi_salfom_2025,
	address = {Cham},
	title = {{SalFoM}: Dynamic Saliency Prediction with Video Foundation Models},
	isbn = {978-3-031-78312-8},
	shorttitle = {{SalFoM}},
	doi = {10.1007/978-3-031-78312-8_3},
	language = {en},
	booktitle = {Pattern Recognition},
	publisher = {Springer Nature Switzerland},
	author = {Moradi, Morteza and Moradi, Mohammad and Rundo, Francesco and Spampinato, Concetto and Borji, Ali and Palazzo, Simone},
	editor = {Antonacopoulos, Apostolos and Chaudhuri, Subhasis and Chellappa, Rama and Liu, Cheng-Lin and Bhattacharya, Saumik and Pal, Umapada},
	year = {2025},
	pages = {33--48},
}

@article{zhou_transformer-based_2023,
	title = {Transformer-Based Multi-Scale Feature Integration Network for Video Saliency Prediction},
	volume = {33},
	issn = {1558-2205},
	url = {https://ieeexplore.ieee.org/document/10130326/authors#authors},
	doi = {10.1109/TCSVT.2023.3278410},
	number = {12},
	urldate = {2025-03-04},
	journal = TCSVT,
	author = {Zhou, Xiaofei and Wu, Songhe and Shi, Ran and Zheng, Bolun and Wang, Shuai and Yin, Haibing and Zhang, Jiyong and Yan, Chenggang},
	month = dec,
	year = {2023},
	Mynote = {Conf.Name: IEEE Transactions on Circuits and Systems for Video Technology},
	pages = {7696--7707},
}

@misc{moradi_transformer-based_2024,
	title = {Transformer-Based Video Saliency Prediction with High Temporal Dimension Decoding},
	url = {http://arxiv.org/abs/2401.07942},
	doi = {10.48550/arXiv.2401.07942},
	urldate = {2025-08-08},
	publisher = {arXiv},
	author = {Moradi, Morteza and Palazzo, Simone and Spampinato, Concetto},
	month = jan,
	year = {2024},
	note = {arXiv:2401.07942},
}

@article{li_tm2sp_2025,
	title = {{TM2SP}: A Transformer-Based Multi-Level Spatiotemporal Feature Pyramid Network for Video Saliency Prediction},
	volume = {35},
	issn = {1558-2205},
	shorttitle = {{TM2SP}},
	url = {https://ieeexplore.ieee.org/abstract/document/10841372/authors},
	doi = {10.1109/TCSVT.2025.3529473},
	number = {6},
	urldate = {2025-08-14},
	journal = TCSVT,
	author = {Li, Chenming and Liu, Shiguang},
	month = jun,
	year = {2025},
	pages = {5236--5250},
}

@article{guo_novel_2010,
	title = {A Novel Multiresolution Spatiotemporal Saliency Detection Model and Its Applications in Image and Video Compression},
	volume = {19},
	issn = {1941-0042},
	url = {https://ieeexplore.ieee.org/document/5223506},
	doi = {10.1109/TIP.2009.2030969},
	number = {1},
	urldate = {2025-08-14},
	journal = TIP,
	author = {Guo, Chenlei and Zhang, Liming},
	year = {2010},
	pages = {185--198},
}

@article{seo_static_2009,
	title = {Static and Space-Time Visual Saliency Detection by Self-Resemblance},
	volume = {9},
	issn = {1534-7362},
	url = {https://doi.org/10.1167/9.12.15},
	doi = {10.1167/9.12.15},
	number = {12},
	urldate = {2025-08-14},
	journal = JOV,
	author = {Seo, Hae Jong and Milanfar, Peyman},
	month = nov,
	year = {2009},
	pages = {15},
}

@inproceedings{hou_dynamic_2008,
	title = {Dynamic Visual Attention: Searching for Coding Length Increments},
	volume = {21},
	shorttitle = {Dynamic Visual Attention},
	url = {https://papers.nips.cc/paper_files/paper/2008/hash/a8baa56554f96369ab93e4f3bb068c22-Abstract.html},
	urldate = {2025-08-14},
	booktitle = NIPS,
	publisher = {Curran Associates, Inc.},
	author = {Hou, Xiaodi and Zhang, Liqing},
	year = {2008},
}

@article{fang_video_2014,
	title = {Video Saliency Incorporating Spatiotemporal Cues and Uncertainty Weighting},
	volume = {23},
	issn = {1941-0042},
	url = {https://ieeexplore.ieee.org/document/6857361},
	doi = {10.1109/TIP.2014.2336549},
	number = {9},
	urldate = {2025-08-14},
	journal = TIP,
	author = {Fang, Yuming and Wang, Zhou and Lin, Weisi and Fang, Zhijun},
	month = sep,
	year = {2014},
	pages = {3910--3921},
}

@inproceedings{khatoonabadi_how_2015,
	title = {How Many Bits Does It Take for a Stimulus to Be Salient?},
	url = {https://ieeexplore.ieee.org/document/7299189},
	doi = {10.1109/CVPR.2015.7299189},
	urldate = {2025-08-14},
	booktitle = CVPR,
	author = {Khatoonabadi, Sayed Hossein and Vasconcelos, Nuno and Bajić, Ivan V. and Shan, Yufeng},
	month = jun,
	year = {2015},
	Mynote = {ISSN: 1063-6919},
	pages = {5501--5510},
}

@article{leboran_dynamic_2017,
	title = {Dynamic Whitening Saliency},
	volume = {39},
	issn = {1939-3539},
	url = {https://ieeexplore.ieee.org/document/7469361},
	doi = {10.1109/TPAMI.2016.2567391},
	number = {5},
	urldate = {2025-08-14},
	journal = PAMI,
	author = {Leborán, Víctor and García-Díaz, Antón and Fdez-Vidal, Xosé R. and Pardo, Xosé M.},
	month = may,
	year = {2017},
	pages = {893--907},
}

@inproceedings{gao_discriminant_2007,
	title = {The Discriminant Center-Surround Hypothesis for Bottom-Up Saliency},
	volume = {20},
	url = {https://papers.nips.cc/paper_files/paper/2007/hash/51ef186e18dc00c2d31982567235c559-Abstract.html},
	urldate = {2025-08-14},
	booktitle = NIPS,
	publisher = {Curran Associates, Inc.},
	author = {Gao, Dashan and Mahadevan, Vijay and Vasconcelos, Nuno},
	year = {2007},
}

@article{mahadevan_spatiotemporal_2010,
	title = {Spatiotemporal Saliency in Dynamic Scenes},
	volume = {32},
	issn = {1939-3539},
	url = {https://ieeexplore.ieee.org/document/4967608},
	doi = {10.1109/TPAMI.2009.112},
	number = {1},
	urldate = {2025-08-14},
	journal = PAMI,
	author = {Mahadevan, Vijay and Vasconcelos, Nuno},
	month = jan,
	year = {2010},
	pages = {171--177},
}

@article{bak_spatio-temporal_2018,
	title = {Spatio-Temporal Saliency Networks for Dynamic Saliency Prediction},
	volume = {20},
	issn = {1941-0077},
	url = {https://ieeexplore.ieee.org/document/8119879},
	doi = {10.1109/TMM.2017.2777665},
	number = {7},
	urldate = {2025-08-14},
	journal = TMM,
	author = {Bak, Cagdas and Kocak, Aysun and Erdem, Erkut and Erdem, Aykut},
	month = jul,
	year = {2018},
	pages = {1688--1698},
}

@inproceedings{gorji_going_2018,
	title = {Going From Image to Video Saliency: Augmenting Image Salience with Dynamic Attentional Push},
	shorttitle = {Going From Image to Video Saliency},
	url = {https://ieeexplore.ieee.org/document/8578881},
	doi = {10.1109/CVPR.2018.00783},
	urldate = {2025-08-14},
	booktitle = CVPR,
	author = {Gorji, Siavash and Clark, James J.},
	month = jun,
	year = {2018},
	Mynote = {ISSN: 2575-7075},
	pages = {7501--7511},
}

@inproceedings{simonyan_two-stream_2014,
	address = {Cambridge, MA, USA},
	series = {{NIPS}'14},
	title = {Two-Stream Convolutional Networks for Action Recognition in Videos},
	volume = {1},
	urldate = {2025-08-13},
	booktitle = NIPS,
	publisher = {MIT Press},
	author = {Simonyan, Karen and Zisserman, Andrew},
	month = dec,
	year = {2014},
	pages = {568--576},
}

@inproceedings{ma_new_2001,
	title = {A New Perceived Motion Based Shot Content Representation},
	volume = {3},
	url = {https://ieeexplore.ieee.org/document/958142},
	doi = {10.1109/ICIP.2001.958142},
	urldate = {2025-08-14},
	booktitle = ICIP,
	author = {Ma, Yu-Fei and Zhang, Hong-Jiang},
	month = oct,
	year = {2001},
	pages = {426--429},
}

@inproceedings{wu_salsac_2020,
	title = {Salsac: A Video Saliency Prediction Model With Shuffled Attentions and Correlation-Based ConvLSTM},
	volume = {34},
	booktitle = AAAI,
	author = {Wu, Xinyi and Wu, Zhenyao and Zhang, Jinglin and Ju, Lili and Wang, Song},
	year = {2020},
	Mynote = {Issue: 07},
	pages = {12410--12417},
}

@inproceedings{min_tased-net_2019,
	title = {Tased-Net: Temporally-Aggregating Spatial Encoder-Decoder Network for Video Saliency Detection},
	booktitle = CVPR,
	author = {Min, Kyle and Corso, Jason J},
	year = {2019},
	pages = {2394--2403},
}

@article{lai_video_2019,
	title = {Video Saliency Prediction Using Spatiotemporal Residual Attentive Networks},
	volume = {29},
	journal = TIP,
	author = {Lai, Qiuxia and Wang, Wenguan and Sun, Hanqiu and Shen, Jianbing},
	year = {2019},
	Mynote = {Publisher: IEEE},
	pages = {1113--1126},
}

@article{linardos_simple_2019,
	title = {Simple vs Complex Temporal Recurrences for Video Saliency Prediction},
	journal = {arXiv preprint arXiv:1907.01869},
	author = {Linardos, Panagiotis and Mohedano, Eva and Nieto, Juan Jose and O'Connor, Noel E and Giró-i-Nieto, Xavier and McGuinness, Kevin},
	year = {2019},
}

@inproceedings{wang_revisiting_2018,
	title = {Revisiting Video Saliency: A Large-Scale Benchmark and a New Model},
	booktitle = CVPR,
	author = {Wang, Wenguan and Shen, Jianbing and Guo, Fang and Cheng, Ming-Ming and Borji, Ali},
	year = {2018},
	pages = {4894--4903},
}

@article{muthuswamy_salient_2013,
	title = {Salient Motion Detection in Compressed Domain},
	volume = {20},
	number = {10},
	journal = SPL,
	author = {Muthuswamy, Karthik and Rajan, Deepu},
	year = {2013},
	Mynote = {Publisher: IEEE},
	pages = {996--999},
}

@inproceedings{jiang_deepvs_2018,
	title = {Deepvs: A Deep Learning Based Video Saliency Prediction Approach},
	booktitle = ECCV,
	author = {Jiang, Lai and Xu, Mai and Liu, Tie and Qiao, Minglang and Wang, Zulin},
	year = {2018},
	pages = {602--617},
}

@article{ma_video_2022,
	title = {Video Saliency Forecasting Transformer},
	volume = {32},
	copyright = {https://ieeexplore.ieee.org/Xplorehelp/downloads/license-information/IEEE.html},
	issn = {1051-8215, 1558-2205},
	url = {https://ieeexplore.ieee.org/document/9770033/},
	doi = {10.1109/TCSVT.2022.3172971},
	language = {en},
	number = {10},
	urldate = {2025-08-14},
	journal = TCSVT,
	author = {Ma, Cheng and Sun, Haowen and Rao, Yongming and Zhou, Jie and Lu, Jiwen},
	month = oct,
	year = {2022},
	pages = {6850--6862},
}

@article{wang_spatio-temporal_2023,
	title = {Spatio-Temporal Self-Attention Network for Video Saliency Prediction},
	volume = {25},
	copyright = {https://ieeexplore.ieee.org/Xplorehelp/downloads/license-information/IEEE.html},
	issn = {1520-9210, 1941-0077},
	url = {https://ieeexplore.ieee.org/document/9667292/},
	doi = {10.1109/TMM.2021.3139743},
	language = {en},
	urldate = {2025-08-14},
	journal = TMM,
	author = {Wang, Ziqiang and Liu, Zhi and Li, Gongyang and Wang, Yang and Zhang, Tianhong and Xu, Lihua and Wang, Jijun},
	year = {2023},
	pages = {1161--1174},
}

@inproceedings{jain_vinet_2021,
	address = {Prague, Czech Republic},
	title = {{ViNet}: Pushing the Limits of Visual Modality for Audio-Visual Saliency Prediction},
	copyright = {https://ieeexplore.ieee.org/Xplorehelp/downloads/license-information/IEEE.html},
	isbn = {978-1-6654-1714-3},
	shorttitle = {{ViNet}},
	url = {https://ieeexplore.ieee.org/document/9635989/},
	doi = {10.1109/IROS51168.2021.9635989},
	language = {en},
	urldate = {2025-08-14},
	booktitle = {2021 {IEEE}/{RSJ} {Int.} {Conf.} {Intell.} {Robots} {Syst.} ({IROS})},
	publisher = {IEEE},
	author = {Jain, Samyak and Yarlagadda, Pradeep and Jyoti, Shreyank and Karthik, Shyamgopal and Subramanian, Ramanathan and Gandhi, Vineet},
	month = sep,
	year = {2021},
	pages = {3520--3527},
}

@article{bellitto_hierarchical_2021,
	title = {Hierarchical Domain-Adapted Feature Learning for Video Saliency Prediction},
	volume = {129},
	issn = {0920-5691, 1573-1405},
	url = {https://link.springer.com/10.1007/s11263-021-01519-y},
	doi = {10.1007/s11263-021-01519-y},
	language = {en},
	number = {12},
	urldate = {2025-08-14},
	journal = IJCV,
	author = {Bellitto, G. and Proietto Salanitri, F. and Palazzo, S. and Rundo, F. and Giordano, D. and Spampinato, C.},
	month = dec,
	year = {2021},
	pages = {3216--3232},
}

@inproceedings{droste_unified_2020,
	address = {Cham},
	title = {Unified Image and Video Saliency Modeling},
	volume = {12350},
	isbn = {978-3-030-58557-0 978-3-030-58558-7},
	url = {https://link.springer.com/10.1007/978-3-030-58558-7_25},
	doi = {10.1007/978-3-030-58558-7_25},
	language = {en},
	urldate = {2025-08-14},
	booktitle = ECCV,
	publisher = {Springer Int. Publishing},
	author = {Droste, Richard and Jiao, Jianbo and Noble, J. Alison},
	editor = {Vedaldi, Andrea and Bischof, Horst and Brox, Thomas and Frahm, Jan-Michael},
	year = {2020},
	pages = {419--435},
}

@inproceedings{sinha_region--interest_2004,
	address = {Montreal, Que., Canada},
	title = {Region-of-Interest Based Compressed Domain Video Transcoding Scheme},
	volume = {3},
	isbn = {978-0-7803-8484-2},
	url = {http://ieeexplore.ieee.org/document/1326506/},
	doi = {10.1109/ICASSP.2004.1326506},
	language = {en},
	urldate = {2025-08-14},
	booktitle = ICASSP,
	publisher = {IEEE},
	author = {Sinha, A. and Agarwal, G. and Anbu, A.},
	year = {2004},
	pages = {iii--161--4},
}

@misc{chang_temporal-spatial_2021,
	title = {Temporal-Spatial Feature Pyramid for Video Saliency Detection},
	url = {http://arxiv.org/abs/2105.04213},
	doi = {10.48550/arXiv.2105.04213},
	urldate = {2025-08-14},
	publisher = {arXiv},
	author = {Chang, Qinyao and Zhu, Shiping},
	month = sep,
	year = {2021},
	note = {arXiv:2105.04213 },
}

@article{peters_applying_2008,
	title = {Applying Computational Tools to Predict Gaze Direction in Interactive Visual Environments},
	volume = {5},
	issn = {1544-3558},
	url = {https://doi.org/10.1145/1279920.1279923},
	doi = {10.1145/1279920.1279923},
	number = {2},
	urldate = {2025-08-18},
	journal = {ACM Trans. Appl. Percept.},
	author = {Peters, Robert J. and Itti, Laurent},
	month = may,
	year = {2008},
	pages = {9:1--9:19},
}

@article{hassenstein_systemtheoretische_1956,
	title = {Systemtheoretische Analyse Der Zeit-, Reihenfolgen- Und Vorzeichenauswertung Bei Der Bewegungsperzeption Des Rüsselkäfers Chlorophanus},
	volume = {11},
	copyright = {De Gruyter expressly reserves the right to use all content for commercial text and data mining within the meaning of Section 44b of the German Copyright Act.},
	issn = {1865-7117},
	url = {https://www.degruyterbrill.com/document/doi/10.1515/znb-1956-9-1004/html},
	doi = {10.1515/znb-1956-9-1004},
	language = {en},
	number = {9-10},
	urldate = {2025-08-20},
	journal = {Zeitschrift für Naturforschung B},
	author = {Hassenstein, B. and Reichardt, W.},
	month = oct,
	year = {1956},
	Mynote = {Publisher: De Gruyter},
	pages = {513--524},
}

@inproceedings{varghese_yolov8_2024,
	title = {{YOLOv8}: A Novel Object Detection Algorithm With Enhanced Performance and Robustness},
	shorttitle = {{YOLOv8}},
	url = {https://ieeexplore.ieee.org/document/10533619},
	doi = {10.1109/ADICS58448.2024.10533619},
	urldate = {2025-08-26},
	booktitle = {2024 {Int.} {Conf.} on {Adv.} in {Data} {Eng.} and {Intell.} {Comput.} {Syst.} ({ADICS})},
	author = {Varghese, Rejin and M., Sambath},
	month = apr,
	year = {2024},
	pages = {1--6},
}

@inproceedings{farha_ms-tcn_2019,
	title = {{MS}-{TCN}: Multi-Stage Temporal Convolutional Network for Action Segmentation},
	shorttitle = {{MS}-{TCN}},
	url = {https://openaccess.thecvf.com/content_CVPR_2019/html/Abu_Farha_MS-TCN_Multi-Stage_Temporal_Convolutional_Network_for_Action_Segmentation_CVPR_2019_paper.html},
	urldate = {2025-10-02},
	booktitle = CVPR,
	author = {Farha, Yazan Abu and Gall, Jurgen},
	year = {2019},
	pages = {3575--3584},
}

@inproceedings{carreira_quo_2017,
	title = {Quo Vadis, Action Recognition? A New Model and the Kinetics Dataset},
	shorttitle = {Quo Vadis, Action Recognition?},
	url = {https://openaccess.thecvf.com/content_cvpr_2017/html/Carreira_Quo_Vadis_Action_CVPR_2017_paper.html},
	urldate = {2025-10-02},
	booktitle = PAMI,
	author = {Carreira, Joao and Zisserman, Andrew},
	year = {2017},
	pages = {6299--6308},
}

@inproceedings{chan_anticipating_2017,
	address = {Cham},
	title = {Anticipating Accidents in Dashcam Videos},
	isbn = {978-3-319-54190-7},
	booktitle = ACCV,
	publisher = {Springer Int. Publishing},
	author = {Chan, Fu-Hsiang and Chen, Yu-Ting and Xiang, Yu and Sun, Min},
	editor = {Lai, Shang-Hong and Lepetit, Vincent and Nishino, Ko and Sato, Yoichi},
	year = {2017},
	pages = {136--153},
}

@inproceedings{you_traffic_2020,
	address = {Cham},
	title = {Traffic Accident Benchmark for Causality Recognition},
	isbn = {978-3-030-58571-6},
	booktitle = ECCV,
	publisher = {Springer Int. Publishing},
	author = {You, Tackgeun and Han, Bohyung},
	editor = {Vedaldi, Andrea and Bischof, Horst and Brox, Thomas and Frahm, Jan-Michael},
	year = {2020},
	pages = {540--556},
}

@inproceedings{fang_dada-2000_2019,
	address = {Auckland, New Zealand},
	title = {{DADA}-2000: Can Driving Accident Be Predicted by Driver Attentionƒ Analyzed by A Benchmark},
	url = {https://doi.org/10.1109/ITSC.2019.8917218},
	doi = {10.1109/ITSC.2019.8917218},
	booktitle = {2019 {IEEE} {Intell.} {Transp.} {Syst.} {Conf.} ({ITSC})},
	publisher = {IEEE Press},
	author = {Fang, Jianwu and Yan, Dingxin and Qiao, Jiahuan and Xue, Jianru and Wang, He and Li, Sen},
	year = {2019},
	pages = {4303--4309},
}

@inproceedings{kopuklu_driver_2021,
	address = {Waikoloa, HI, USA},
	title = {Driver Anomaly Detection: A Dataset and Contrastive Learning Approach},
	copyright = {https://doi.org/10.15223/policy-029},
	isbn = {978-1-6654-0477-8},
	shorttitle = {Driver Anomaly Detection},
	url = {https://ieeexplore.ieee.org/document/9423242/},
	doi = {10.1109/WACV48630.2021.00014},
	language = {en},
	urldate = {2025-10-01},
	booktitle = {2021 {IEEE} {Winter} {Conf.} on {Appl.} of {Comput.} {Vis..} ({WACV})},
	publisher = {IEEE},
	author = {Kopuklu, Okan and Zheng, Jiapeng and Xu, Hang and Rigoll, Gerhard},
	month = jan,
	year = {2021},
	pages = {91--100},
}

@book{noauthor_global_2023,
	address = {Geneva},
	edition = {1st ed},
	title = {Global Status Report on Road Safety 2023},
	isbn = {978-92-4-008651-7},
	language = {en},
	publisher = {World Health Organization},
	year = {2023},
}

@article{tian_designing_2023,
	title = {Designing {BERT} for Convolutional Networks: Sparse and Hierarchical Masked Modeling},
	journal = {arXiv:2301.03580},
	author = {Tian, Keyu and Jiang, Yi and Diao, Qishuai and Lin, Chen and Wang, Liwei and Yuan, Zehuan},
	year = {2023},
}

@article{zhang_sun_2008,
	title = {{SUN}: A Bayesian Framework for Saliency Using Natural Statistics},
	volume = {8},
	issn = {1534-7362},
	shorttitle = {{SUN}},
	url = {http://jov.arvojournals.org/article.aspx?doi=10.1167/8.7.32},
	doi = {10.1167/8.7.32},
	language = {en},
	number = {7},
	urldate = {2024-07-22},
	journal = JOV,
	author = {Zhang, Lingyun and Tong, Matthew H. and Marks, Tim K. and Shan, Honghao and Cottrell, Garrison W.},
	month = dec,
	year = {2008},
	pages = {32},
}

@article{kim_fast_2018,
	title = {Fast 2D Complex Gabor Filter With Kernel Decomposition},
	volume = {27},
	copyright = {https://ieeexplore.ieee.org/Xplorehelp/downloads/license-information/IEEE.html},
	issn = {1057-7149, 1941-0042},
	url = {http://ieeexplore.ieee.org/document/8207611/},
	doi = {10.1109/TIP.2017.2783621},
	language = {en},
	number = {4},
	urldate = {2024-09-18},
	journal = TIP,
	author = {Kim, Jaeyoon and Um, Suhyuk and Min, Dongbo},
	month = apr,
	year = {2018},
	pages = {1713--1722},
}

@article{le_meur_predicting_2007,
	title = {Predicting Visual Fixations on Video Based on Low-Level Visual Features},
	volume = {47},
	copyright = {https://www.elsevier.com/tdm/userlicense/1.0/},
	issn = {00426989},
	url = {https://linkinghub.elsevier.com/retrieve/pii/S0042698907002593},
	doi = {10.1016/j.visres.2007.06.015},
	language = {en},
	number = {19},
	urldate = {2025-02-11},
	journal = VR,
	author = {Le Meur, Olivier and Le Callet, Patrick and Barba, Dominique},
	month = sep,
	year = {2007},
	pages = {2483--2498},
}

@inproceedings{sun_scalability_2020,
	title = {Scalability in Perception for Autonomous Driving: Waymo Open Dataset},
	shorttitle = {Scalability in Perception for Autonomous Driving},
	url = {https://ieeexplore.ieee.org/document/9156973},
	doi = {10.1109/CVPR42600.2020.00252},
	urldate = {2025-10-29},
	booktitle = CVPR,
	author = {Sun, Pei and Kretzschmar, Henrik and Dotiwalla, Xerxes and Chouard, Aurélien and Patnaik, Vijaysai and Tsui, Paul and Guo, James and Zhou, Yin and Chai, Yuning and Caine, Benjamin and Vasudevan, Vijay and Han, Wei and Ngiam, Jiquan and Zhao, Hang and Timofeev, Aleksei and Ettinger, Scott and Krivokon, Maxim and Gao, Amy and Joshi, Aditya and Zhang, Yu and Shlens, Jonathon and Chen, Zhifeng and Anguelov, Dragomir},
	month = jun,
	year = {2020},
	Mynote = {ISSN: 2575-7075},
	pages = {2443--2451},
}

@inproceedings{ettinger_large_2021,
	title = {Large Scale Interactive Motion Forecasting for Autonomous Driving : The Waymo Open Motion Dataset},
	shorttitle = {Large Scale Interactive Motion Forecasting for Autonomous Driving},
	url = {https://ieeexplore.ieee.org/document/9709630},
	doi = {10.1109/ICCV48922.2021.00957},
	urldate = {2025-10-29},
	booktitle = ICCV,
	author = {Ettinger, Scott and Cheng, Shuyang and Caine, Benjamin and Liu, Chenxi and Zhao, Hang and Pradhan, Sabeek and Chai, Yuning and Sapp, Ben and Qi, Charles and Zhou, Yin and Yang, Zoey and Chouard, Aurélien and Sun, Pei and Ngiam, Jiquan and Vasudevan, Vijay and McCauley, Alexander and Shlens, Jonathon and Anguelov, Dragomir},
	month = oct,
	year = {2021},
	Mynote = {ISSN: 2380-7504},
	pages = {9690--9699},
}

@inproceedings{codevilla_exploring_2019,
	title = {Exploring the Limitations of Behavior Cloning for Autonomous Driving},
	url = {https://ieeexplore.ieee.org/document/9009463},
	doi = {10.1109/ICCV.2019.00942},
	urldate = {2025-10-29},
	booktitle = ICCV,
	author = {Codevilla, Felipe and Santana, Eder and Lopez, Antonio and Gaidon, Adrien},
	month = oct,
	year = {2019},
	Mynote = {ISSN: 2380-7504},
	pages = {9328--9337},
}

@misc{zimmer_safety-critical_2025,
	title = {Safety-Critical Learning for Long-Tail Events: The {TUM} Traffic Accident Dataset},
	shorttitle = {Safety-Critical Learning for Long-Tail Events},
	url = {http://arxiv.org/abs/2508.14567},
	doi = {10.48550/arXiv.2508.14567},
	urldate = {2025-10-29},
	publisher = {arXiv},
	author = {Zimmer, Walter and Greer, Ross and Zhou, Xingcheng and Song, Rui and Pavel, Marc and Lehmberg, Daniel and Ghita, Ahmed and Gopalkrishnan, Akshay and Trivedi, Mohan and Knoll, Alois},
	month = aug,
	year = {2025},
	Mynote = {arXiv:2508.14567 [cs]},
}

@inproceedings{bao_uncertainty-based_2020,
	address = {New York, NY, USA},
	series = {{MM} '20},
	title = {Uncertainty-Based Traffic Accident Anticipation With Spatio-Temporal Relational Learning},
	isbn = {978-1-4503-7988-5},
	url = {https://dl.acm.org/doi/10.1145/3394171.3413827},
	doi = {10.1145/3394171.3413827},
	urldate = {2025-10-29},
	booktitle = ACMMM,
	publisher = {Association for Computing Machinery},
	author = {Bao, Wentao and Yu, Qi and Kong, Yu},
	month = oct,
	year = {2020},
	pages = {2682--2690},
}

@article{yao_dota_2023,
	title = {{DoTA}: Unsupervised Detection of Traffic Anomaly in Driving Videos},
	volume = {45},
	issn = {1939-3539},
	shorttitle = {{DoTA}},
	doi = {10.1109/TPAMI.2022.3150763},
	language = {eng},
	number = {1},
	journal = PAMI,
	author = {Yao, Yu and Wang, Xizi and Xu, Mingze and Pu, Zelin and Wang, Yuchen and Atkins, Ella and Crandall, David J.},
	month = jan,
	year = {2023},
	pmid = {35157576},
	pages = {444--459},
}

@article{hadizadeh_eye-tracking_2012,
	title = {Eye-Tracking Database for a Set of Standard Video Sequences},
	volume = {21},
	issn = {1941-0042},
	url = {https://ieeexplore.ieee.org/document/5986709},
	doi = {10.1109/TIP.2011.2165292},
	number = {2},
	urldate = {2025-10-29},
	journal = TIP,
	author = {Hadizadeh, Hadi and Enriquez, Mario J. and Bajic, Ivan V.},
	month = feb,
	year = {2012},
	pages = {898--903},
}

@article{mathe_actions_2015,
	title = {Actions in the Eye: Dynamic Gaze Datasets and Learnt Saliency Models for Visual Recognition},
	volume = {37},
	issn = {1939-3539},
	shorttitle = {Actions in the Eye},
	url = {https://ieeexplore.ieee.org/document/6942210},
	doi = {10.1109/TPAMI.2014.2366154},
	number = {7},
	urldate = {2025-10-29},
	journal = PAMI,
	author = {Mathe, Stefan and Sminchisescu, Cristian},
	month = jul,
	year = {2015},
	pages = {1408--1424},
}

@article{mital_clustering_2011,
	title = {Clustering of Gaze During Dynamic Scene Viewing Is Predicted by Motion},
	volume = {3},
	issn = {1866-9964},
	url = {https://doi.org/10.1007/s12559-010-9074-z},
	doi = {10.1007/s12559-010-9074-z},
	language = {en},
	number = {1},
	urldate = {2025-10-29},
	journal = {{Cogn.} {Comput.}},
	author = {Mital, Parag K. and Smith, Tim J. and Hill, Robin L. and Henderson, John M.},
	month = mar,
	year = {2011},
	pages = {5--24},
}

@inproceedings{ma_model_2002,
	title = {A Model of Motion Attention for Video Skimming},
	volume = {1},
	url = {https://ieeexplore.ieee.org/document/1037976},
	doi = {10.1109/ICIP.2002.1037976},
	urldate = {2025-11-03},
	booktitle = ICIP,
	author = {Ma, Yu-Fei and Zhang, Hong-Jiang},
	month = sep,
	year = {2002},
	Mynote = {ISSN: 1522-4880},
	pages = {I--I},
}

@inproceedings{guo_spatio-temporal_2008,
	title = {Spatio-Temporal Saliency Detection Using Phase Spectrum of Quaternion Fourier Transform},
	url = {https://ieeexplore.ieee.org/document/4587715},
	doi = {10.1109/CVPR.2008.4587715},
	urldate = {2025-11-03},
	booktitle = CVPR,
	author = {Guo, Chenlei and Ma, Qi and Zhang, Liming},
	month = jun,
	year = {2008},
	Mynote = {ISSN: 1063-6919},
	pages = {1--8},
}

@article{kim_spatiotemporal_2011,
	title = {Spatiotemporal Saliency Detection and Its Applications in Static and Dynamic Scenes},
	volume = {21},
	issn = {1558-2205},
	url = {https://ieeexplore.ieee.org/document/5728853},
	doi = {10.1109/TCSVT.2011.2125450},
	number = {4},
	urldate = {2025-11-03},
	journal = TCSVT,
	author = {Kim, Wonjun and Jung, Chanho and Kim, Changick},
	month = apr,
	year = {2011},
	pages = {446--456},
}

@article{fang_deep3dsaliency_2018,
	title = {{Deep3DSaliency}: Deep Stereoscopic Video Saliency Detection Model by 3D Convolutional Networks},
	issn = {1941-0042},
	shorttitle = {{Deep3DSaliency}},
	doi = {10.1109/TIP.2018.2885229},
	language = {eng},
	journal = TIP,
	author = {Fang, Yuming and Ding, Guanqun and Li, Jia and Fang, Zhijun},
	month = dec,
	year = {2018},
}

@inproceedings{wildes_measure_1998,
	title = {A Measure of Motion Salience for Surveillance Applications},
	url = {https://ieeexplore.ieee.org/document/727163},
	doi = {10.1109/ICIP.1998.727163},
	urldate = {2025-11-04},
	booktitle = ICIP,
	author = {Wildes, R.P.},
	month = oct,
	year = {1998},
	pages = {183--187 vol.3},
}

@article{wixson_detecting_2000,
	title = {Detecting Salient Motion by Accumulating Directionally-Consistent Flow},
	volume = {22},
	issn = {1939-3539},
	url = {https://ieeexplore.ieee.org/document/868680},
	doi = {10.1109/34.868680},
	number = {8},
	urldate = {2025-11-04},
	journal = PAMI,
	author = {Wixson, L.},
	month = aug,
	year = {2000},
	pages = {774--780},
}

@article{itti_bayesian_2009,
	series = {Visual {Attention}: {Psychophysics}, electrophysiology and neuroimaging},
	title = {Bayesian Surprise Attracts Human Attention},
	volume = {49},
	issn = {0042-6989},
	url = {https://www.sciencedirect.com/science/article/pii/S0042698908004380},
	doi = {10.1016/j.visres.2008.09.007},
	number = {10},
	urldate = {2025-11-04},
	journal = VR,
	author = {Itti, Laurent and Baldi, Pierre},
	month = jun,
	year = {2009},
	pages = {1295--1306},
}

@article{ma_generic_2005,
	title = {A Generic Framework of User Attention Model and Its Application in Video Summarization},
	volume = {7},
	issn = {1941-0077},
	url = {https://ieeexplore.ieee.org/document/1510638},
	doi = {10.1109/TMM.2005.854410},
	number = {5},
	urldate = {2025-11-04},
	journal = TMM,
	author = {Ma, Yu-Fei and Hua, Xian-Sheng and Lu, Lie and Zhang, Hong-Jiang},
	month = oct,
	year = {2005},
	pages = {907--919},
}

@inproceedings{zhai_visual_2006,
	address = {New York, NY, USA},
	series = {{MM} '06},
	title = {Visual Attention Detection in Video Sequences Using Spatiotemporal Cues},
	isbn = {978-1-59593-447-5},
	url = {https://doi.org/10.1145/1180639.1180824},
	doi = {10.1145/1180639.1180824},
	urldate = {2025-11-04},
	booktitle = ACMMM,
	publisher = {Association for Computing Machinery},
	author = {Zhai, Yun and Shah, Mubarak},
	month = oct,
	year = {2006},
	pages = {815--824},
}

@article{li_traffic_2025,
	title = {Traffic Accident Risk Prediction Based on Deep Learning and Spatiotemporal Features of Vehicle Trajectories},
	volume = {20},
	issn = {1932-6203},
	url = {https://journals.plos.org/plosone/article?id=10.1371/journal.pone.0320656},
	doi = {10.1371/journal.pone.0320656},
	language = {en},
	number = {5},
	urldate = {2025-11-05},
	journal = {PLOS ONE},
	author = {Li, Hao and Chen, Linbing},
	month = may,
	year = {2025},
	Mynote = {Publisher: Public Library of Science},
	pages = {e0320656},
}

@article{li_prediction_2025,
	title = {Prediction of Traffic Accident Risk Based on Vehicle Trajectory Data},
	volume = {26},
	issn = {1538-957X},
	doi = {10.1080/15389588.2024.2402936},
	language = {eng},
	number = {2},
	journal = {Traffic Injury Prevention},
	author = {Li, Hao and Yu, Lina},
	year = {2025},
	pmid = {39570198},
	pages = {164--171},
}

@article{cornia_predicting_2018,
	title = {Predicting Human Eye Fixations via an {LSTM}-Based Saliency Attentive Model},
	volume = {27},
	issn = {1941-0042},
	url = {https://ieeexplore.ieee.org/document/8400593},
	doi = {10.1109/TIP.2018.2851672},
	number = {10},
	urldate = {2025-11-06},
	journal = TIP,
	author = {Cornia, Marcella and Baraldi, Lorenzo and Serra, Giuseppe and Cucchiara, Rita},
	month = oct,
	year = {2018},
	pages = {5142--5154},
}

@article{zhang_video_2019,
	title = {Video Saliency Prediction Based on Spatial-Temporal Two-Stream Network},
	volume = {29},
	issn = {1558-2205},
	url = {https://ieeexplore.ieee.org/document/8543830},
	doi = {10.1109/TCSVT.2018.2883305},
	number = {12},
	urldate = {2025-11-06},
	journal = TCSVT,
	author = {Zhang, Kao and Chen, Zhenzhong},
	month = dec,
	year = {2019},
	pages = {3544--3557},
}

@article{zhang_spatial-temporal_2021,
	title = {A Spatial-Temporal Recurrent Neural Network for Video Saliency Prediction},
	volume = {30},
	issn = {1941-0042},
	url = {https://ieeexplore.ieee.org/document/9263359},
	doi = {10.1109/TIP.2020.3036749},
	urldate = {2025-11-06},
	journal = TIP,
	author = {Zhang, Kao and Chen, Zhenzhong and Liu, Shan},
	year = {2021},
	pages = {572--587},
}

@article{itti_automatic_2004,
	title = {Automatic Foveation for Video Compression Using a Neurobiological Model of Visual Attention},
	volume = {13},
	issn = {1057-7149},
	doi = {10.1109/tip.2004.834657},
	language = {eng},
	number = {10},
	journal = TIP,
	author = {Itti, Laurent},
	month = oct,
	year = {2004},
	pmid = {15462141},
	pages = {1304--1318},
}

@article{ali_advances_2024,
	title = {Advances, Challenges, and Future Research Needs in Machine Learning-Based Crash Prediction Models: A Systematic Review},
	volume = {194},
	issn = {0001-4575},
	shorttitle = {Advances, Challenges, and Future Research Needs in Machine Learning-Based Crash Prediction Models},
	url = {https://www.sciencedirect.com/science/article/pii/S0001457523004256},
	doi = {10.1016/j.aap.2023.107378},
	urldate = {2025-11-07},
	journal = {Accident Analysis \& Prevention},
	author = {Ali, Yasir and Hussain, Fizza and Haque, Md Mazharul},
	month = jan,
	year = {2024},
	pages = {107378},
}

@article{karim_dynamic_2022,
	title = {A Dynamic Spatial-Temporal Attention Network for Early Anticipation of Traffic Accidents},
	volume = {23},
	issn = {1524-9050},
	url = {https://doi.org/10.1109/TITS.2022.3155613},
	doi = {10.1109/TITS.2022.3155613},
	number = {7},
	urldate = {2025-11-07},
	journal = {{IEEE} {Trans.} on {Intell.} {Transp} {Syst.}},
	author = {Karim, Muhammad Monjurul and Li, Yu and Qin, Ruwen and Yin, Zhaozheng},
	month = jul,
	year = {2022},
	pages = {9590--9600},
}

@article{article_noauthor_ycbcr_2025,
  title={Studio Encoding Parameters of Digital Television for Standard 4:3 and Wide-Screen 16:9 Aspect Ratios},
  author={BT, RIR and others},
  journal={Int. radio consultative committee Int. telecommunication union, Switzerland, CCIR Rep},
  year={2011}
}

@inproceedings{BaoICCV2021DRIVE,
  author = "Bao, Wentao and Yu, Qi and Kong, Yu",
  title = "Deep Reinforced Accident Anticipation with Visual Explanation",
  booktitle = "International Conference on Computer Vision (ICCV)",
  year = "2021"
}

@article{OpenImages,
  author = {Alina Kuznetsova and Hassan Rom and Neil Alldrin and Jasper Uijlings and Ivan Krasin and Jordi Pont-Tuset and Shahab Kamali and Stefan Popov and Matteo Malloci and Alexander Kolesnikov and Tom Duerig and Vittorio Ferrari},
  title = {The Open Images Dataset V4: Unified image classification, object detection, and visual relationship detection at scale},
  year = {2020},
  journal = {IJCV}
}

@inproceedings{DBSCAN_1996,
author = {Ester, Martin and Kriegel, Hans-Peter and Sander, J\"{o}rg and Xu, Xiaowei},
title = {A density-based algorithm for discovering clusters in large spatial databases with noise},
year = {1996},
publisher = {AAAI Press},
booktitle = {Proceedings of the Second International Conference on Knowledge Discovery and Data Mining},
pages = {226–231},
numpages = {6},
location = {Portland, Oregon},
series = {KDD'96}
}

@INPROCEEDINGS{Brockmann1999levy,
  author={Brockmann, D. and Geisel, T.},
  booktitle={1999 Ninth International Conference on Artificial Neural Networks ICANN 99. (Conf. Publ. No. 470)}, 
  title={Are human scanpaths Levy flights?}, 
  year={1999},
  volume={1},
  number={},
  pages={263-268 vol.1},
  doi={10.1049/cp:19991119}}

@article{Giuseppe2004Constrained,
title = {Modelling gaze shift as a constrained random walk},
journal = {Physica A: Statistical Mechanics and its Applications},
volume = {331},
number = {1},
pages = {207-218},
year = {2004},
issn = {0378-4371},
doi = {https://doi.org/10.1016/j.physa.2003.09.011},
url = {https://www.sciencedirect.com/science/article/pii/S0378437103008331},
author = {Giuseppe Boccignone and Mario Ferraro}
}

@article{marlow2015temporal,
  title={Temporal structure of human gaze dynamics is invariant during free viewing},
  author={Marlow, Colleen A and Viskontas, Indre V and Matlin, Alisa and Boydston, Cooper and Boxer, Adam and Taylor, Richard P},
  journal={PloS one},
  volume={10},
  number={9},
  pages={e0139379},
  year={2015},
  publisher={Public Library of Science San Francisco, CA USA}
}




\vfill

\end{document}